\documentclass{article}
\usepackage{iftex}
\ifPDFTeX
  \usepackage[utf8]{inputenc}
\fi
\usepackage[T1]{fontenc}
\usepackage{genai4health_2026}
\usepackage{hyperref}
\usepackage{url}
\usepackage{booktabs}
\usepackage{amsmath,amssymb}
\usepackage{graphicx}
\hypersetup{hidelinks}

\usepackage{comment}
\usepackage{subcaption}
\usepackage{multirow}
\usepackage{array}
\usepackage{amsfonts}
\usepackage{nicefrac}
\usepackage{microtype}
\usepackage{xcolor}
\newcommand{\cbox}[1]{\textcolor{#1}{\rule{1.6ex}{1.6ex}}}
\usepackage{pifont}

\usepackage{float}
\usepackage{placeins}
\usepackage{enumitem}
\usepackage{tikz}
\usetikzlibrary{arrows.meta,shapes.geometric,positioning,fit,backgrounds,calc}
\usepackage{pgfplots}
\pgfplotsset{compat=1.18}
\pgfplotsset{
  barlegend/.style={
    legend image code/.code={\fill[#1] (0cm,-0.1cm) rectangle (0.3cm,0.1cm);},
  },
}
\usepgfplotslibrary{groupplots}

\definecolor{vizone}{HTML}{2a78d6}
\definecolor{viztwo}{HTML}{eb6834}
\definecolor{vizthree}{HTML}{1baf7a}
\definecolor{vizmauve}{HTML}{9c5a8c}
\definecolor{vizred}{HTML}{d1495b}

\title{Adversarial Attacks Expose Path-Dependent Failures in Discrete Medical Image Tokenizers}
\author{%
  Ahmed Loughzali\thanks{Equal contribution.} \\
  IMT Nord Europe \\
  \texttt{ahmed.loughzali@etu.imt-nord-europe.fr}
  \And
  Akash Yallamati\footnotemark[1] \\
  University of South Florida \\
  \texttt{akash34@usf.edu}
  \And
  Aadiv Reki \\
  Purdue University \\
  \texttt{areki@purdue.edu}
  \And
  Aslan Shaukat Vaida \\
  Nanyang Technological University \\
  \texttt{vaid0005@e.ntu.edu.sg}
  \And
  Jeyashree Krishnan \\
  Algoverse AI Research \\
  \texttt{jeyashree.krishnan@pm.me}
}

\begin{document}
\maketitle

\begin{abstract}
{Discrete image tokenizers convert medical images into token sequences that downstream models either decode back into an image for a classifier (the reconstruction path), classify directly (the token path), or feed to a generative multimodal model \citep{ma2025meditok}.
We ask whether an imperceptible perturbation of the input image can break these systems through the tokenizer alone.
Building on general-domain tokenizer red-teaming \citep{bhagwatkar2026adversarial}, we evaluate four tokenizers on three imaging modalities and extend the strongest attack to a deployed generative model.
Our central finding is \emph{path-dependence}: the same tokenizer fails in opposite ways depending on how its output is used.
A supervised attack collapses diagnostic accuracy on the reconstruction path, with area under the ROC curve (AUC) falling to $\leq 0.013$ at $\epsilon=2/255$ on all four tokenizers (0.5 is chance), while sparing the token path.
An unsupervised, label-free attack does the reverse; its damage is concentrated in the multi-codebook tokenizers and is worst in MedITok, whose token-path AUC falls from $0.919$ to $0.239$.
The same label-free attack transfers to LLaVA-Med, a frozen medical vision-language model whose image encoder is MedITok, making it report proliferative diabetic retinopathy in a healthy retina or call a malignant lesion benign, without access to the question, the answer, or the language model's gradients.
Harm is driven by how far the attack moves the tokenizer's internal representation, not by how many tokens change: on the learned-codebook tokenizers, a control attack that changes nearly all tokens but moves the representation less leaves every path and answer intact.
In our comparison, domain-specific pretraining does not provide robustness: MedITok has the best clean accuracy yet is the most fragile, more so than UniTok, its general-domain counterpart with the same architecture.
None of six candidate defenses repairs every failure mode, and several fail once the attacker optimizes against the defense directly.
Robustness of a medical tokenizer must therefore be evaluated separately for each downstream path, against such an adaptive adversary.}
\end{abstract}

\section{Introduction}
\label{sec:introduction}

\begin{figure}[t]
\centering
\includegraphics[width=0.85\linewidth]{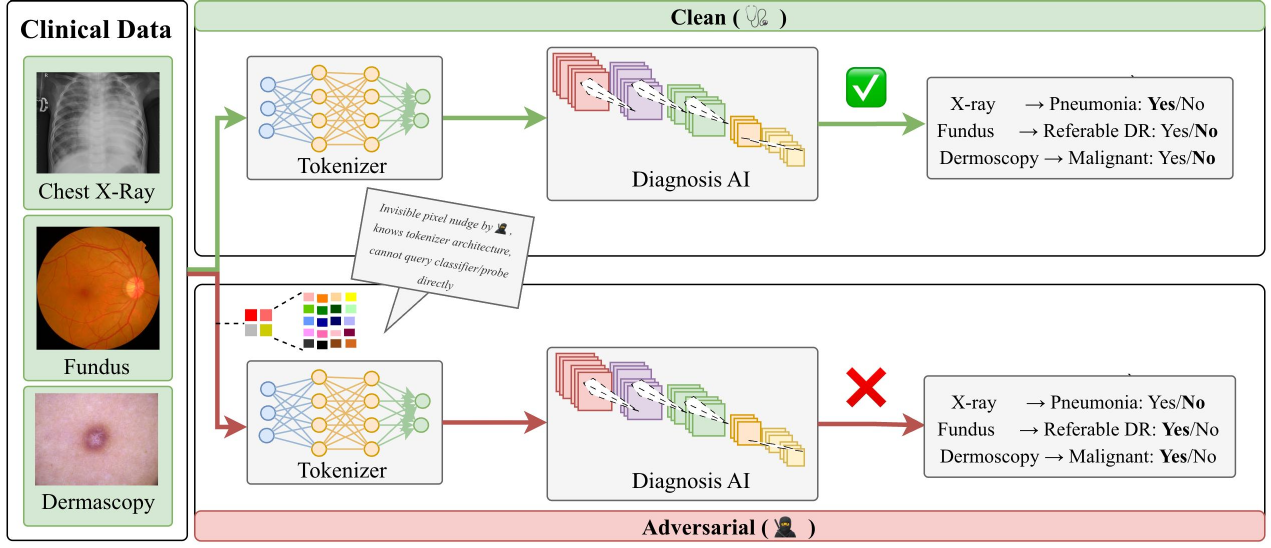}

\caption{Threat model. An adversary who knows the tokenizer architecture but cannot
query the downstream model adds an imperceptible pixel-space perturbation to a
clinical image. The same tokenizer yields the correct diagnosis for the clean image
and the wrong one for the perturbed image, across classification and generative
downstream models alike.}
\label{fig:threat-model}

\end{figure}

Medical imaging is adopting artificial intelligence rapidly, increasingly through
generative and multimodal models that read a scan and produce a diagnosis, a report,
or an answer to a clinical question \citep{bannur2023learning,zhang2023biomedclip}.
Many of these systems, including several built for medical imaging, share one
upstream component: a discrete image tokenizer that converts an image into a short
sequence of discrete codes \citep{ma2025meditok,sun2024autoregressive}. The
downstream model consumes these codes in place of pixels, in one of three ways: (a)
decoding them back into an image before classifying it (the reconstruction path), (b)
classifying the discrete codes directly (the token path), or (c) feeding them to a
generative multimodal model that answers questions about the scan (the generative path).

Because the tokenizer sits upstream of all of these, a vulnerability in it can propagate
to the models built on it. Medical AI's susceptibility to adversarial perturbations is
well established. Small, imperceptible pixel changes flip predictions across
classification, detection, and segmentation systems and modalities
\citep{finlayson2019adversarial,madry2018towards,kaviani2022adversarial,paschali2018generalizability}. But existing attacks target the downstream model directly;
the tokenizer, now upstream of many and largely open-source with weights and
codebooks freely inspectable \citep{ma2025meditok,sun2024autoregressive,esser2021taming},
is a distinct and largely unstudied surface (Figure~\ref{fig:threat-model}).

We red-team this surface across all three downstream paths, evaluating four
tokenizers (LlamaGen VQ-16, MedITok, UniTok, Finite Scalar Quantization (FSQ)) over
three modalities (chest radiography, fundus photography, dermoscopy). Our central
finding, path-dependence, is that the same imperceptible perturbation breaks a
tokenizer in opposite ways depending on how its output is used: a task-supervised
attack collapses the reconstruction path while sparing the token path, and an
unsupervised, label-free attack that pushes the tokenizer's internal features across the
boundaries where its token assignments change (a boundary-directed attack) does the reverse. This second
attack, computed from the public tokenizer weights alone, propagates to the
generative path: it makes a frozen medical vision-language model (LLaVA-Med, vision
encoder MedITok) assert that a healthy retina shows proliferative diabetic
retinopathy, or that a malignant lesion is benign, without ever seeing the question,
the answer, or a gradient from the language model. Our boundary-directed attacks
build on the isotropic embedding-displacement attack of
\citet{bhagwatkar2026adversarial}: we target the quantization boundary directly and
separate crossing it from displacing past it. Our primary contribution is not the attack
objective itself but the path-dependence analysis it enables across tokenizers,
modalities, and downstream uses.

Our contributions are as follows.

\begin{itemize}

  \item We establish path-dependence as the organizing property of tokenizer
    robustness: supervised and unsupervised attacks collapse opposite downstream
    paths, so no single number characterizes a tokenizer and robustness must be
    measured per path. To our knowledge, this is the first systematic adversarial evaluation of
    discrete tokenizers in the medical domain.
  \item We red-team a deployed generative medical model (LLaVA-Med) through its
    tokenizer, showing a label-free, imperceptible perturbation makes it invent or
    erase disease with no access to the language model or the task.
  \item We isolate displacement past the quantization boundary, not the token flip,
    as the cause of harm: a near-$100\%$ flip rate does not predict diagnostic risk.
  \item In our comparison, domain-specific pretraining does not provide robustness: MedITok is the most
    clean-accurate tokenizer and the most fragile under attack.
  \item We evaluate six defenses and find none repairs every failure mode; the gap
    survives adaptive attack \citep{athalye2018obfuscated}, and only stochastic
    quantization smoothing carries a formal guarantee \citep{cohen2019certified}.

\end{itemize}

\section{Related work}
\label{sec:related}
%%%%%%%%%%%%%%%%%%%%%%%%%%%%%%%%%%%%%%%%%%%%%%%%%%%%%%%%%%%%

\textbf{Adversarial robustness in medical imaging.}
Adversarial examples are small, designed perturbations that cause misclassification \citep{szegedy2014intriguing,goodfellow2015explaining}, typically optimized via projected gradient descent (PGD) \citep{madry2018towards,carlini2017evaluating,croce2020reliable}.
In medical imaging, imperceptible perturbations cause diagnostic failure in chest-radiograph, retinal, and histopathology classifiers \citep{finlayson2019adversarial}. More aggressive attacks like CT-GAN insert or remove visible disease signs from volumetric scans via generative models
\citep{mirsky2019ctgan}. These are content-level manipulations that produce qualitatively different failure modes from the bounded, pixel-space attacks we study.

\textbf{Discrete image tokenization.} Vector-quantized models replace continuous encoder outputs with entries from a finite codebook \citep{oord2017neural,esser2021taming}. Later work improves compression, uses multiple codebooks, or replaces learned codebooks with scalar quantization \citep{yu2024titok,ma2025unitok,mentzer2024finite}, and medical variants like MedITok extend these to clinical tasks \citep{ma2025meditok}. Despite their increasing use, the adversarial failure modes introduced by the quantization stage remain underexplored.

\textbf{Robustness of tokenized representations.} Prior work has shown that
adversarial perturbations disrupt latent representations and autoencoder
reconstructions \citep{gondim2018adversarial,kos2018adversarial}, but on continuous
encoders; the discrete quantization step has received almost no attention. Closest to ours, \citet{bhagwatkar2026adversarial} attack general-purpose discrete tokenizers by maximizing pre-quantization embedding displacement and show that the resulting failure modes transfer across classification, retrieval, and captioning, including targeted attacks that steer a multimodal model's generated caption toward an attacker-chosen, sometimes unsafe, string. This is red-teaming of generative
multimodal systems through the tokenizer, but on general-domain models and imagery.
We extend this line to the medical domain and to a deployed medical vision-language
model, and target the quantization boundary directly rather than maximizing isotropic
displacement: bounded and unbounded objectives separate crossing a boundary from
displacing past it, motivated by prior work showing that distance to the nearest
decision boundary predicts local vulnerability
\citep{fawzi2017analysis,moosavi2016deepfool}.

\textbf{Defending against adversarial attacks.} Certified robustness gives a
mathematical guarantee via randomized smoothing \citep{cohen2019certified}, while many
defenses instead rely on obfuscated gradients that stop a naive but not an adaptive
attacker \citep{athalye2018obfuscated}, which we test via Backward Pass Differentiable Approximation (BPDA). Closest to our defense
setting, \citet{keepitreal2025} show a compression-based purification's adaptive
resistance depends on decoder realism, motivating our decoder fine-tune's GAN-based
objective.

\section{Methods}
\label{sec:methodology}
%%%%%%%%%%%%%%%%%%%%%%%%%%%%%%%%%%%%%%%%%%%%%%%%%%%%%%%%%%%%

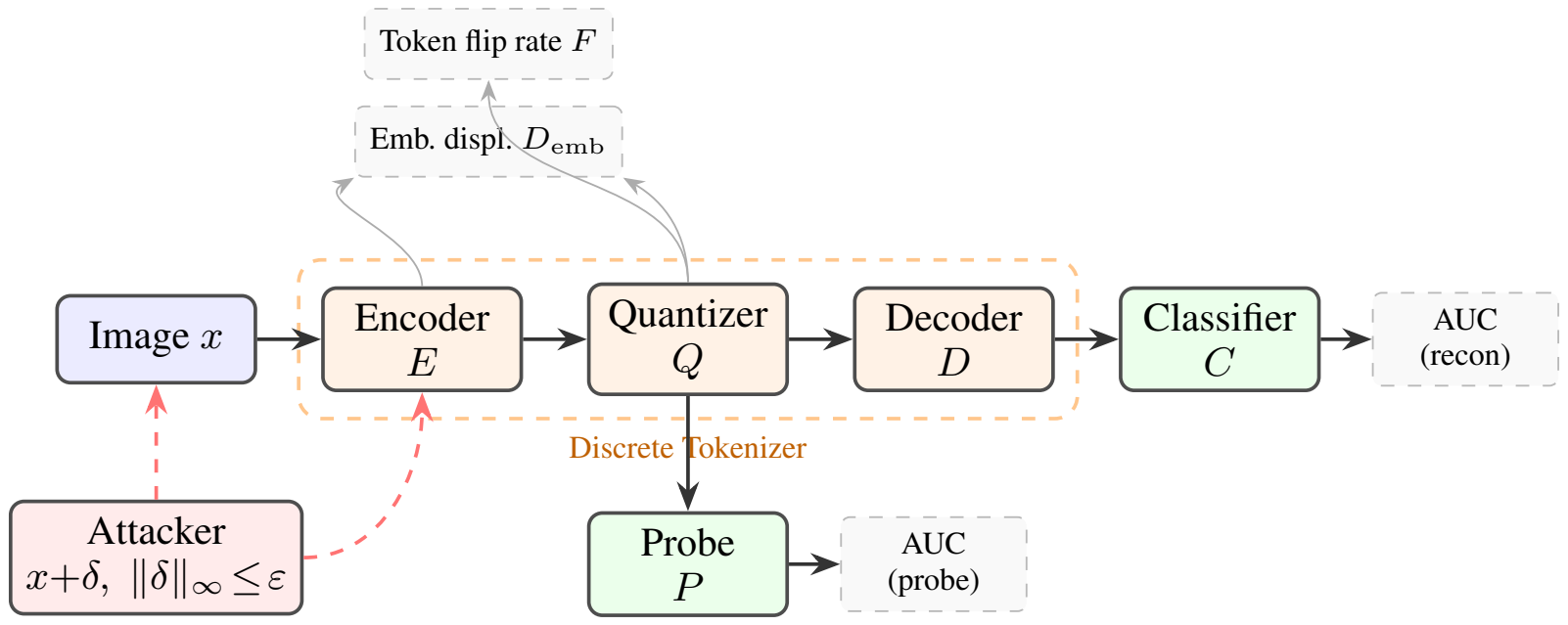
\begin{figure}[htbp]
\centering
\resizebox{0.75\linewidth}{!}{%
\begin{tikzpicture}[
    node distance=0.5cm and 0.5cm,
    pipe/.style={
        rectangle, rounded corners=3pt,
        minimum width=1.55cm, minimum height=0.68cm,
        align=center, font=\small,
        draw=black!70, thick
    },
    imgbox/.style={pipe, fill=blue!8},
    atkbox/.style={pipe, fill=red!8},
    tokbox/.style={pipe, fill=orange!10},
    clsbox/.style={pipe, fill=green!8},
    metbox/.style={
        rectangle, rounded corners=2pt,
        minimum width=1.45cm, minimum height=0.55cm,
        align=center, font=\scriptsize,
        draw=gray!55, fill=gray!5, dashed
    },
    arr/.style={-Stealth, thick, black!80},
    advarr/.style={-Stealth, thick, dashed, red!55},
    metarr/.style={-Stealth, thin, gray!65},
]

%% ---- Main pipeline (single row) ----
\node[imgbox] (img)   {Image $x$};
\node[tokbox,  right=0.5cm of img]   (enc)   {Encoder\\$E$};
\node[tokbox,  right=0.5cm of enc]   (quant) {Quantizer\\$Q$};
\node[tokbox,  right=0.5cm of quant] (dec)   {Decoder\\$D$};
\node[clsbox,  right=0.5cm of dec]   (clf)   {Classifier\\$C$};
\node[metbox,  right=0.4cm of clf]   (aucr)  {AUC\\(recon)};

\draw[arr] (img)   -- (enc);
\draw[arr] (enc)   -- (quant);
\draw[arr] (quant) -- (dec);
\draw[arr] (dec)   -- (clf);
\draw[arr] (clf)   -- (aucr);

%% ---- Probe branch (below quantizer) ----
\node[clsbox, below=0.9cm of quant] (probe) {Probe\\$P$};
\node[metbox, right=0.4cm of probe] (aucp)  {AUC\\(probe)};

\draw[arr] (quant.south) -- (probe.north);
\draw[arr] (probe) -- (aucp);

%% ---- Attacker (below image) ----
\node[atkbox, below=0.9cm of img] (atk)
    {Attacker\\$x{+}\delta,\;\|\delta\|_\infty\!\leq\!\varepsilon$};
\draw[advarr] (atk.north) -- (img.south);
\draw[advarr] (atk.east)  to[out=0,in=270] (enc.south);

%% ---- Metric: Embedding displacement (above, between enc and quant) ----
\node[metbox, above=0.85cm of enc, xshift=0.52cm] (emb)
    {Emb.\ displ.\ $D_{\mathrm{emb}}$};
\draw[metarr] (enc.north)   to[out=90,in=210] (emb.south west);
\draw[metarr] (quant.north) to[out=90,in=330] (emb.south east);

%% ---- Metric: Token flip (above quantizer) ----
\node[metbox, above=0.2cm of emb] (flip) {Token flip rate $F$};
\draw[metarr] (quant.north) to[out=90,in=270] (flip.south);

%% ---- Tokenizer bounding box ----
\begin{pgfonlayer}{background}
  \node[draw=orange!45, dashed, rounded corners=5pt, thick,
        fit=(enc)(quant)(dec),
        inner sep=5pt,
        label={[font=\scriptsize, orange!75!black]below:Discrete Tokenizer}
  ] {};
\end{pgfonlayer}

\end{tikzpicture}%
}
\caption{Evaluation pipeline. A bounded pixel-space
perturbation enters the tokenizer. The resulting codes are consumed
along a reconstruction path ($D\!\to\!C$) or a probe path ($P$ on
post-quantization features). Metrics are measured at the marked stages.}
\label{fig:pipeline-overview}
\end{figure}

Figure~\ref{fig:pipeline-overview} shows our evaluation pipeline. A tokenizer's output is consumed in two ways: discrete codes are decoded back into pixels before classification (reconstruction path), or a classifier reads the post-quantization features directly (token path). The two paths are mechanistically independent at the quantizer boundary, so an attack can collapse one while leaving the other intact. We evaluate both paths simultaneously and instrument the quantizer with token flip rate and embedding displacement to localize where each attack causes harm.

\FloatBarrier

%%%%%%%%%%%%%%%%%%%%%%%%%%%%%%%%%%%%%%%%%%%%%%%%%%%%%%%%%%%%
\subsection{Setup}
\label{sec:setup}
%%%%%%%%%%%%%%%%%%%%%%%%%%%%%%%%%%%%%%%%%%%%%%%%%%%%%%%%%%%%

Let $x$ be an input image. A discrete tokenizer consists of an encoder $E$,
quantizer $Q$, and decoder $D$: the encoder maps $x$ to continuous latents
$z=E(x)$, the quantizer assigns each latent a discrete code $q=Q(z)$ (a quantization boundary
is where this assignment changes, so a latent pushed across one flips its token), and the
decoder reconstructs $\hat{x}=D(q)$ for a downstream classifier. A linear
probe operating on $q$ directly, with no decoder, lets us separate damage at
the tokenizer level from damage that survives reconstruction. An adversary
constructs $x_{\mathrm{adv}} = x + \delta$ with $\|\delta\|_\infty \leq
\epsilon$.

We evaluate four tokenizers (LlamaGen VQ-16, FSQ, UniTok, MedITok) spanning the main quantization strategies, and three medical imaging datasets (Kermany chest X-ray, APTOS fundus photography, ISIC dermoscopy) covering different modalities and classifier architectures. Tables, full dataset details, and hyperparameters are in Appendix~\ref{app:models-datasets} and Appendix~\ref{app:hyperparams}. We report three metrics per image: token flip rate, embedding displacement, and downstream diagnostic area under the ROC curve (AUC); the certified radius metric applies only to Stochastic Quantization Smoothing (SQS), and reconstruction fidelity (LPIPS/PSNR/SSIM) applies only to the reconstruction path, both introduced in Section~\ref{sec:defenses} and formally defined in Appendix~\ref{app:metrics-detail}.

%%%%%%%%%%%%%%%%%%%%%%%%%%%%%%%%%%%%%%%%%%%%%%%%%%%%%%%%%%%%
\subsection{Attack methods}
\label{sec:attacks}
%%%%%%%%%%%%%%%%%%%%%%%%%%%%%%%%%%%%%%%%%%%%%%%%%%%%%%%%%%%%

Our attacks differ in what information the adversary needs and divide into two classes:

\begin{itemize}[noitemsep, topsep=2pt]
  \item \textbf{Supervised (M0, M1):} require the diagnostic label and backpropagate a task loss through the classifier, differing only in whether the gradient also routes through the tokenizer.
  \item \textbf{Unsupervised (M2):} use no labels and operate purely on the tokenizer's geometry, specifically how far a latent sits from its nearest quantization boundary, available to any adversary holding just the public tokenizer weights.
\end{itemize}

All attacks use PGD \citep{madry2018towards} under the same $\ell_\infty$ budget, differing only in objective.

\begin{itemize}

  \item \textbf{M0, direct classifier attack (reference baseline).} The adversary
  optimizes the perturbation directly against the classifier loss, bypassing the tokenizer. This is a reference confirming the downstream classifier is itself vulnerable.

  \item \textbf{M1, tokenizer-mediated supervised attack.} The adversary optimizes
  the classifier loss through the full tokenizer pipeline, so the
  perturbation is shaped by the encoder, quantizer, and decoder. This is our supervised, task-aware attack.

  \item \textbf{M2-unb, boundary-directed geometric attack (unbounded).} Without labels, the adversary maximizes a margin objective driving each latent towards its nearest competing quantization boundary and past it. This is our unsupervised, quantization-aware attack. Unlike prior displacement objectives \citep{bhagwatkar2026adversarial}, which maximize pre-quantization displacement isotropically, ours targets codebook geometry directly, though it degenerates toward pure displacement once tokens flip, a point we revisit in Section~\ref{sec:experiments}.

  \item \textbf{M2-sat, boundary-directed geometric attack (saturated).} Identical to the unbounded variant until a token flips, at which point the objective stops
  rewarding further movement. It therefore flips a comparable fraction of tokens but does not push latents further once they cross a boundary, separating the effect of a token flip from that of continued displacement. This serves as a mechanistic control for the unbounded attack.

\end{itemize}

\begin{table}[t]
\centering
\footnotesize
\setlength{\tabcolsep}{4pt}
\caption{Attack settings and adversary access. All attacks use PGD under the same $\ell_\infty$ budget.}
\label{tab:attack-settings}
\begin{tabular}{l l l >{\raggedright\arraybackslash}p{3.1cm} >{\raggedright\arraybackslash}p{3.3cm} >{\raggedright\arraybackslash}p{1.9cm}}
\toprule
Attack & Label & Classifier grads & Gradient path & Objective & Role \\
\midrule
M0 & Required & Required & Classifier only (bypasses tokenizer) & Classifier loss & Reference baseline \\
M1 & Required & Required & Encoder, quantizer, decoder, classifier & Classifier loss & Supervised attack \\
M2-unb & Not needed & Not needed & Tokenizer only (public weights) & Margin toward and past the nearest boundary & Unsupervised attack \\
M2-sat & Not needed & Not needed & Tokenizer only (public weights) & Same margin, stops rewarding once a token flips & Control for M2-unb \\
\bottomrule
\end{tabular}
\end{table}

FSQ has no learned codebook and therefore no Voronoi boundaries. Its margin is measured against the nearest scalar rounding threshold, so its geometric attack is an architecture-specific adaptation diverging from other tokenizers.

%%%%%%%%%%%%%%%%%%%%%%%%%%%%%%%%%%%%%%%%%%%%%%%%%%%%%%%%%%%%
\subsection{Defense methods}
\label{sec:defense-methods}
%%%%%%%%%%%%%%%%%%%%%%%%%%%%%%%%%%%%%%%%%%%%%%%%%%%%%%%%%%%%

We evaluate six defenses spanning three approaches: noise-based (Pixel-noise, SQS), encoder fine-tuning (Encoder-FT, Encoder-FT+SQS), and decoder fine-tuning (Decoder-FT, Encoder-FT+Decoder-FT). Each is also tested for robustness under adaptive attack via Backward Pass Differentiable Approximation (BPDA) and a held-out Carlini-Wagner attack.

\begin{itemize}

  \item \textbf{Pixel-noise.} Additive Gaussian noise applied directly to the input pixels before the tokenizer.

  \item \textbf{Stochastic Quantization Smoothing (SQS).} Noise injected at the quantization step, then majority-voted over repeated re-quantizations \citep{cohen2019certified}, with hyperparameters swept per tokenizer.

  \item \textbf{Encoder-FT.} Encoder-only fine-tuning against a boundary-margin loss on adversarial examples, pushing clean and adversarial inputs toward the same code. Adapts \citet{bhagwatkar2026adversarial}'s unsupervised fine-tuning defense to our boundary-margin objective. Combined with SQS as Encoder-FT+SQS.

  \item \textbf{Decoder-FT.} Decoder-only fine-tuning with a GAN, perceptual, and reconstruction loss on clean and adversarial codes, so reconstructions stay plausible even from attacker codes. Realistic reconstructions resist adaptive purification better than pixel-distortion minimizing reconstructions \citep{keepitreal2025}. Combined with Encoder-FT as Encoder-FT+Decoder-FT.

  \item \textbf{BPDA.} A straight-through gradient estimate through the quantization step, attacking the defended model directly rather than reusing a perturbation from the undefended tokenizer \citep{athalye2018obfuscated}. This tests whether a defense's apparent robustness survives an attacker who knows the defense exists.

  \item \textbf{Held-out Carlini-Wagner $\ell_2$ attack.} A second adaptive check with a different objective and norm than any attack the defenses were tuned against \citep{carlini2017evaluating}.

\end{itemize}

\section{Results}
\label{sec:results}
%%%%%%%%%%%%%%%%%%%%%%%%%%%%%%%%%%%%%%%%%%%%%%%%%%%%%%%%%%%%

We evaluate four attacks across four tokenizers and three medical imaging datasets,
extend the strongest to a deployed generative model, and test whether any of six
defenses closes both failure modes.

\subsection{Attack results}
\label{sec:experiments}

The reference attack M0 (perturbing classifier directly, no tokenizer)
drives AUC to $0.00$ at every budget on every dataset, confirming the

downstream classifier alone is vulnerable. All numbers below use the $\ell_\infty$ budget on the $[-1,1]$ normalized input, averaged over the three datasets at $\epsilon=2/255$ unless stated otherwise.

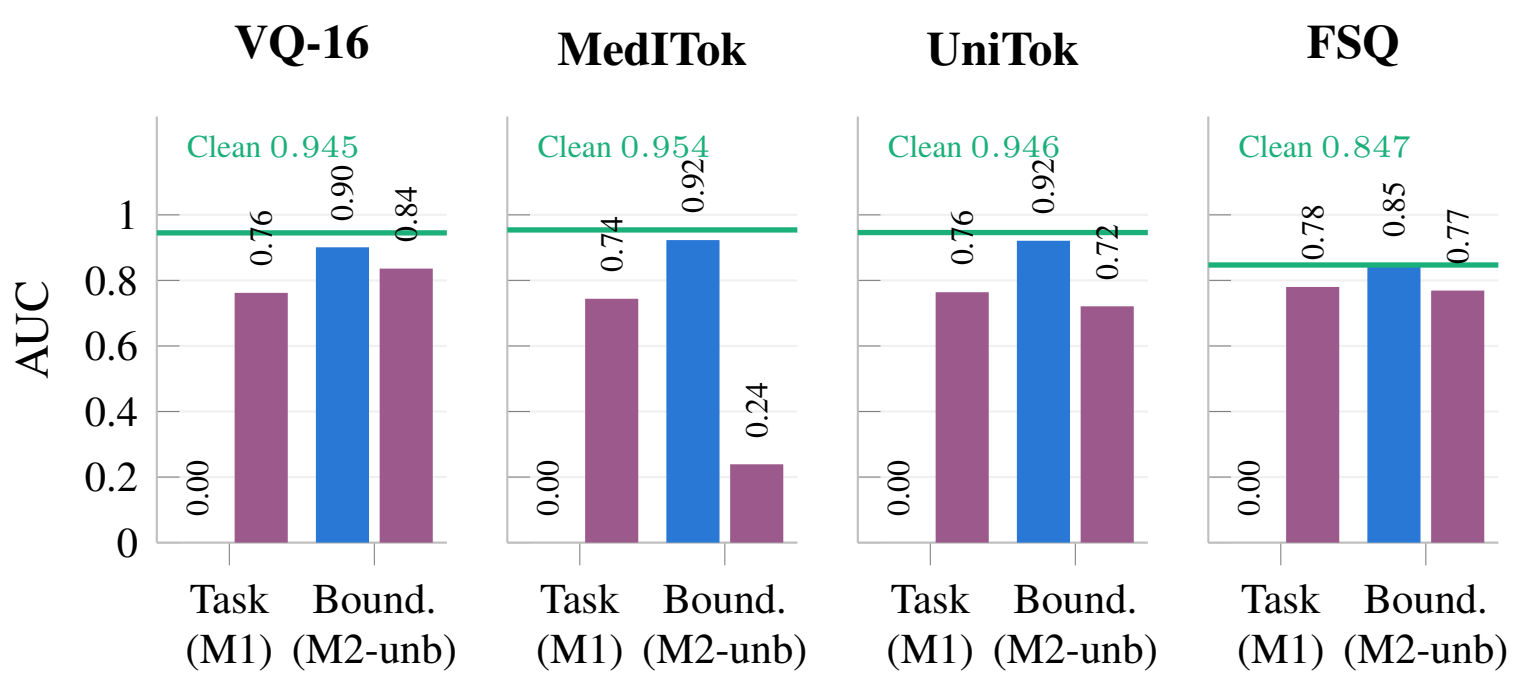
\begin{figure}[htbp]
\centering
\begin{tikzpicture}
\begin{groupplot}[
    group style={group size=4 by 1, horizontal sep=0.4cm,
        ylabels at=edge left, yticklabels at=edge left},
    width=3.5cm, height=4.4cm,
    ymin=0, ymax=1.3, ytick={0,0.2,0.4,0.6,0.8,1}, yticklabels={0,0.2,0.4,0.6,0.8,1},
    symbolic x coords={M1,M2-unb}, xtick=data,
    xticklabels={Task\\(M1), Bound.\\(M2-unb)},
    tick label style={font=\footnotesize, align=center}, label style={font=\small},
    title style={font=\small\bfseries},
    ymajorgrids, grid style={gray!12},
    axis lines*=left, axis line style={gray!50},
    enlarge x limits=0.5, ybar,
    every axis/.append style={bar width=10pt},
    every axis plot/.append style={draw=none},
    nodes near coords, point meta=explicit symbolic,
    nodes near coords style={font=\tiny, rotate=90, anchor=west, xshift=1.5pt,
        /pgf/number format/fixed, /pgf/number format/precision=2,
        /pgf/number format/zerofill},
    ]

\nextgroupplot[title={VQ-16}, ylabel={AUC}]
\addplot[fill=vizone] coordinates {(M1,0.0)[0.00] (M2-unb,0.901)[0.90]};
\addplot[fill=vizmauve] coordinates {(M1,0.762)[0.76] (M2-unb,0.836)[0.84]};
\draw[vizthree, line width=1pt]
    ({rel axis cs:0,0}|-{axis cs:M1,0.945}) -- ({rel axis cs:1,0}|-{axis cs:M1,0.945});
\node[anchor=north west, font=\tiny, text=vizthree] at (rel axis cs:0.04,1.0) {Clean $0.945$};

\nextgroupplot[title={MedITok}]
\addplot[fill=vizone] coordinates {(M1,0.000)[0.00] (M2-unb,0.923)[0.92]};
\addplot[fill=vizmauve] coordinates {(M1,0.744)[0.74] (M2-unb,0.239)[0.24]};
\draw[vizthree, line width=1pt]
    ({rel axis cs:0,0}|-{axis cs:M1,0.954}) -- ({rel axis cs:1,0}|-{axis cs:M1,0.954});
\node[anchor=north west, font=\tiny, text=vizthree] at (rel axis cs:0.04,1.0) {Clean $0.954$};

\nextgroupplot[title={UniTok}]
\addplot[fill=vizone] coordinates {(M1,0.000)[0.00] (M2-unb,0.921)[0.92]};
\addplot[fill=vizmauve] coordinates {(M1,0.764)[0.76] (M2-unb,0.721)[0.72]};
\draw[vizthree, line width=1pt]
    ({rel axis cs:0,0}|-{axis cs:M1,0.946}) -- ({rel axis cs:1,0}|-{axis cs:M1,0.946});
\node[anchor=north west, font=\tiny, text=vizthree] at (rel axis cs:0.04,1.0) {Clean $0.946$};

\nextgroupplot[title={FSQ}]
\addplot[fill=vizone] coordinates {(M1,0.000)[0.00] (M2-unb,0.850)[0.85]};
\addplot[fill=vizmauve] coordinates {(M1,0.780)[0.78] (M2-unb,0.769)[0.77]};
\draw[vizthree, line width=1pt]
    ({rel axis cs:0,0}|-{axis cs:M1,0.847}) -- ({rel axis cs:1,0}|-{axis cs:M1,0.847});
\node[anchor=north west, font=\tiny, text=vizthree] at (rel axis cs:0.04,1.0) {Clean $0.847$};

\end{groupplot}
\end{tikzpicture}
\caption{Path-dependence. Mean adversarial AUC on the reconstruction
(\cbox{vizone}) and probe (\cbox{vizmauve}) paths. The green line marks each
tokenizer's clean baseline (value labeled top-left of each panel). The two
attacks damage opposite paths.}
\label{fig:spine}
\end{figure}

\begin{table}[htbp]
\centering
\small
\setlength{\tabcolsep}{5pt}
\renewcommand{\arraystretch}{0.88}
\caption{Attack $\times$ inference path at $\epsilon=2/255$, mean over three datasets. Clean is the no-attack AUC. On the learned-codebook tokenizers, M2-sat flips as many tokens as M2-unb yet damages neither path. AUC columns measure attack success ($\downarrow$ = more damage). $n{=}256$ per cell.}
\label{tab:main}
\begin{tabular}{llccc}
\toprule
Tokenizer & Attack & AUC$_{\text{recon}}\downarrow$ & AUC$_{\text{probe}}\downarrow$ & Flip\%$\uparrow$ \\
\midrule
\multirow{4}{*}{LlamaGen VQ-16} & Clean (no attack) & 0.945 & 0.863 & --- \\
 & M1 (supervised) & 0.005 & 0.762 & 85.0 \\
 & M2-sat (geom., bounded)   & 0.940 & 0.864 & 96.7 \\
 & M2-unb (geom., unbounded) & 0.901 & 0.836 & 98.0 \\
\midrule
\multirow{4}{*}{MedITok} & Clean (no attack) & 0.954 & 0.919 & --- \\
 & M1 (supervised) & 0.000 & 0.744 & 81.3 \\
 & M2-sat (geom., bounded)   & 0.951 & 0.868 & 99.3 \\
 & M2-unb (geom., unbounded) & 0.923 & 0.239 & 98.6 \\
\midrule
\multirow{4}{*}{UniTok} & Clean (no attack) & 0.946 & 0.907 & --- \\
 & M1 (supervised) & 0.000 & 0.764 & 83.4 \\
 & M2-sat (geom., bounded)   & 0.939 & 0.886 & 99.0 \\
 & M2-unb (geom., unbounded) & 0.921 & 0.721 & 99.0 \\
\midrule
\multirow{4}{*}{FSQ} & Clean (no attack) & 0.847 & 0.855 & --- \\
 & M1 (supervised) & 0.000 & 0.780 & 43.7 \\
 & M2-sat (geom., bounded)   & 0.842 & 0.764 & 96.5 \\
 & M2-unb (geom., unbounded) & 0.850 & 0.769 & 80.6 \\
\bottomrule
\end{tabular}
\renewcommand{\arraystretch}{1}
\end{table}

The same perturbation breaks a tokenizer differently depending on downstream use (Figure~\ref{fig:spine}, Table~\ref{tab:main}). M1 collapses reconstruction (AUC $\leq 0.013$, all tokenizers) while sparing the probe ($0.74$--$0.78$); M2-unb reverses this, keeping reconstruction usable ($0.85$--$0.92$) while collapsing MedITok's probe ($0.919$ to $0.239$). No single number describes a tokenizer's robustness because it depends on the inference path. The collapse concentrates in multi-codebook tokenizers; VQ-16's probe barely moves (drop $0.028$), failing on reconstruction instead.

This path-split also clarifies what causes the harm. M2-sat and M2-unb flip nearly identical fractions of tokens on learned-codebook tokenizers, both near $100\%$. Only M2-unb, which keeps pushing past the boundary, collapses the probe (MedITok: $0.24$ vs. $0.87$ for M2-sat, at matched flip rates). At matched flips, it moves the embedding roughly twice as far, on every tokenizer and modality except FSQ (Appendix~\ref{app:cross-tokenizer}). Harm tracks displacement past the boundary, not the flip itself, and a $100\%$ flip rate says nothing about diagnostic risk.

The tokenizer architecture shapes how badly each path fails. MedITok and UniTok share an architecture but differ in pretraining data; other training differences may also contribute, so this comparison is suggestive rather than causal. In it, medical pretraining does not provide robustness. MedITok has the best clean probe AUC ($0.919$) yet the worst M2-unb collapse (drop $0.680$, $74\%$ of its clean AUC, vs.\ $0.186$/$0.086$/$0.028$, i.e.\ $21\%$/$10\%$/$3\%$, for UniTok/FSQ/VQ-16), an ordering that holds across all three modalities. The most clinically informative representation is the most attackable.

FSQ stands apart. Its reconstruction still collapses to $0.000$ under M1, so it is
not robust in general; it holds up only against the geometric attack (probe/recon
drops of $0.09$/$0.00$). This reflects its codebook-free design: its margin targets
scalar rounding thresholds rather than Voronoi boundaries, and its clean baseline is
already weak ($0.810$ Kermany, $0.797$ ISIC). Displacement is read within a tokenizer
only ($\approx 1.8$ for FSQ to $75$ for VQ-16 under M2-unb).

\subsection{Generative downstream}
\label{sec:generative}
\begin{figure}[t]
\centering
\footnotesize
\setlength{\tabcolsep}{4pt}
\renewcommand{\arraystretch}{0.95}
\begin{tabular}{@{}m{1.8cm} >{\raggedright\arraybackslash}m{3.2cm} >{\raggedright\arraybackslash}m{3.7cm} >{\raggedright\arraybackslash}m{3.7cm}@{}}
\toprule
\textbf{Input} & \textbf{Clean} &
\textbf{M2-sat} ($\epsilon{=}8/255$) & \textbf{M2-unb} ($\epsilon{=}8/255$) \\
& & {\scriptsize flips $100\%$, disp.\ $38.7$} & {\scriptsize flips $100\%$, disp.\ $71.1$} \\
\midrule

{\setlength{\fboxsep}{0pt}\fbox{\includegraphics[width=1.6cm]{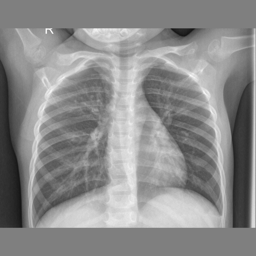}}}
& \emph{Which part of the body?}\par\vspace{2pt}
  \textbf{Thoracic region}, the chest area.
& \textcolor{vizthree}{\textbf{Chest region}, the thoracic cavity.}
& \textcolor{vizred}{\textbf{Abdominal region}, showing the liver.} \\
\midrule

{\setlength{\fboxsep}{0pt}\fbox{\includegraphics[width=1.6cm]{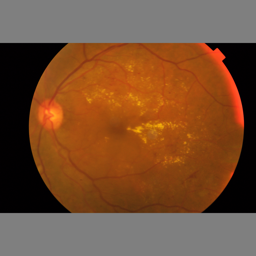}}}
& \emph{Which part of the body?}\par\vspace{2pt}
  The \textbf{retina}, at the back of the eye.
& \textcolor{vizthree}{The \textbf{retina}, at the back of the eye.}
& \textcolor{vizred}{The \textbf{abdominal region}, showing the liver.} \\
\cmidrule(l){2-4}

& \emph{Evidence of diabetic retinopathy?}\par\vspace{2pt}
  \textbf{Yes}: microaneurysms and hemorrhages.
& \textcolor{vizthree}{\textbf{Yes}: microaneurysms and hemorrhages.}
& \textcolor{vizred}{\textbf{No}, the retina appears normal.} \\
\midrule

{\setlength{\fboxsep}{0pt}\fbox{\includegraphics[width=1.6cm]{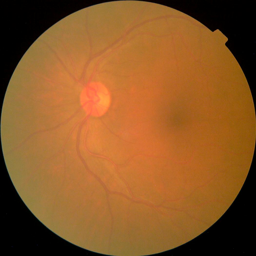}}}
& \emph{Evidence of diabetic retinopathy?}\par\vspace{2pt}
  \textbf{No}, the retinal vessels appear normal.
& \textcolor{vizthree}{\textbf{No}, the retinal vessels are normal.}
& \textcolor{vizred}{\textbf{Yes}, \textbf{proliferative} DR with neovascularization.} \\
\midrule

{\setlength{\fboxsep}{0pt}\fbox{\includegraphics[width=1.6cm]{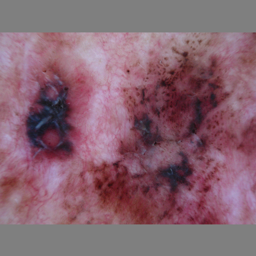}}}
& \emph{Benign or malignant?}\par\vspace{2pt}
  \textbf{Malignant}: irregular, nodular, ulcerated.
& \textcolor{vizthree}{\textbf{Malignant}: irregular, nodular, ulcerated.}
& \textcolor{vizred}{\textbf{Benign}, no signs of malignancy.} \\
\bottomrule
\end{tabular}

\caption{Verbatim answers from a frozen LLaVA-Med (vision encoder MedITok) at
$\epsilon{=}8/255$. \textcolor{vizthree}{Green} answers agree with the clean image;
\textcolor{vizred}{red} answers are corrupted. Rows~3--4 are the same question on
two images of opposite class. $n{=}99$ pooled.}
\label{fig:vqa}
\end{figure}

To test whether the same failure propagates beyond classification, we attack
LLaVA-Med, a multimodal medical visual question answering (VQA) model whose vision encoder is MedITok. We run
the identical label-free M2-unb attack on the input image, pass the perturbed image
through the frozen VQA model, and read its answer (Figure~\ref{fig:vqa}). The attack
corrupts modality and anatomy answers across all three modalities and, on the
clinical questions, breaks in both directions: it erases disease that is present and
invents disease that is absent. As in Section~\ref{sec:experiments}, M2-unb and its
saturated counterpart M2-sat flip the same $\approx100\%$ of tokens, yet M2-unb
displaces the embedding roughly twice as far and destroys nearly every answer
($1\%$ still correct) while M2-sat leaves $98\%$ intact; a matched-$\ell_\infty$
random perturbation, which flips $75\%$ of tokens, is likewise harmless. Harm
tracks displacement past the boundary, not the flip. This is MedITok-only and
inference-only, on a small set of images; full transcripts are in
Appendix~\ref{app:vqa}.

\FloatBarrier

\subsection{Defense results}
\label{sec:defenses}

Section~\ref{sec:experiments} showed that the reconstruction and token paths fail under different attacks. Table~\ref{tab:defense_recon_all} reports reconstruction-path and probe-path AUC under M2-unb at the worst-case budget ($\epsilon=8/255$), averaged over three datasets; full per-dataset results and the Encoder-FT+SQS combination are in Appendix~\ref{app:full-sweeps}.

\begin{table}[htbp]
\centering
\small
\setlength{\tabcolsep}{3.9pt}
\renewcommand{\arraystretch}{0.88}
\caption{Defense comparison, reconstruction-path vs.\ probe-path AUC.
AUC$\uparrow$ under M2-unb, $\epsilon=8/255$, mean over the three datasets
($n{=}128$ each). Clean is the no-attack AUC. None is the
undefended attacked AUC. Encoder-FT and Encoder-FT+Decoder-FT use the encoder
fine-tuned at the attack's own budget.}
\label{tab:defense_recon_all}
\begin{tabular}{llccccccc}
\toprule
Tokenizer & Path & Clean & None & Pixel-noise & SQS & Encoder-FT & Decoder-FT & +Decoder-FT \\
\midrule
\multirow{2}{*}{LlamaGen VQ-16} & Recon & 0.942 & 0.760 & 0.766 & 0.661 & 0.740 & 0.896 & 0.890 \\
 & Probe & 0.871 & 0.728 & 0.738 & 0.724 & 0.756 & 0.728 & 0.756 \\
\midrule
\multirow{2}{*}{MedITok} & Recon & 0.947 & 0.697 & 0.889 & 0.719 & 0.720 & 0.869 & 0.870 \\
 & Probe & 0.913 & 0.104 & 0.828 & 0.105 & 0.125 & 0.104 & 0.125 \\
\midrule
\multirow{2}{*}{UniTok} & Recon & 0.937 & 0.856 & 0.902 & 0.843 & 0.847 & 0.890 & 0.900 \\
 & Probe & 0.903 & 0.409 & 0.835 & 0.410 & 0.653 & 0.409 & 0.653 \\
\midrule
\multirow{2}{*}{FSQ} & Recon & 0.831 & 0.775 & 0.773 & 0.761 & 0.756 & 0.881 & 0.840 \\
 & Probe & 0.786 & 0.596 & 0.597 & 0.605 & 0.630 & 0.596 & 0.630 \\
\bottomrule
\end{tabular}
\end{table}

No single defense closes both paths simultaneously for any tokenizer. The defense behaviors fall into three groups:

\begin{itemize}
    \item \textbf{Token-path and partial reconstruction recovery.} Pixel-noise is the strongest token-path defense, raising MedITok probe AUC from $0.104$ to $0.828$ and UniTok from $0.409$ to $0.835$. For these two tokenizers it also partially recovers the reconstruction path ($0.697\to0.889$ for MedITok; $0.856\to0.902$ for UniTok). For VQ-16 and FSQ, pixel-noise provides negligible recovery on either path.

    \item \textbf{Reconstruction-path recovery only.} Decoder-FT is the strongest reconstruction-path defense (MedITok: $0.697\to0.869$), but leaves probe AUC at the undefended level, since fine-tuning the decoder does not alter the pre-quantization features the probe reads. Encoder-FT+Decoder-FT provides no additional probe-path benefit over Decoder-FT alone.

    \item \textbf{Attack-dependent protection.} SQS retains $0.73$--$0.94$ of clean AUC against the boundary-directed attacks (M2-sat/M2-unb), but collapses to $0.000$ on every tokenizer against the supervised attack M1 (Appendix~\ref{app:full-sweeps}, Figure~\ref{fig:sqs-attack-dependence}). Combining it with Encoder-FT (Encoder-FT+SQS) changes probe AUC by less than $0.013$ relative to Encoder-FT alone.
\end{itemize}

\begin{figure}[htbp]
\centering
\begin{tikzpicture}
\begin{groupplot}[
    group style={group size=4 by 1, horizontal sep=0.4cm,
        ylabels at=edge left, yticklabels at=edge left},
    width=3.6cm, height=4.4cm,
    ymin=0, ymax=1.3, ytick={0,0.2,0.4,0.6,0.8,1}, yticklabels={0,0.2,0.4,0.6,0.8,1},
    symbolic x coords={N,PN,DFT}, xtick=data,
    tick label style={font=\footnotesize}, label style={font=\small},
    title style={font=\small\bfseries},
    ymajorgrids, grid style={gray!12},
    axis lines*=left, axis line style={gray!50},
    enlarge x limits=0.28, ybar,
    every axis/.append style={bar width=7pt},
    every axis plot/.append style={draw=none},
    nodes near coords, point meta=explicit symbolic,
    nodes near coords style={font=\tiny, rotate=90, anchor=west, xshift=1.5pt,
        /pgf/number format/fixed, /pgf/number format/precision=2,
        /pgf/number format/zerofill},
    ]
\nextgroupplot[title={VQ-16}, ylabel={AUC}]
\addplot[fill=vizone, bar shift=-4pt]
    coordinates {(N,0.760)[0.76] (PN,0.766)[0.77] (DFT,0.895)[0.90]};
\addplot[fill=vizmauve, bar shift=4pt]
    coordinates {(N,0.728)[0.73] (PN,0.738)[0.74] (DFT,0.728)[0.73]};
\draw[vizthree, line width=1pt]
    ({rel axis cs:0,0}|-{axis cs:N,0.942}) -- ({rel axis cs:1,0}|-{axis cs:N,0.942});
\node[anchor=north west, font=\tiny, text=vizthree] at (rel axis cs:0.04,1.0) {Clean $0.94$};

\nextgroupplot[title={MedITok}]
\addplot[fill=vizone, bar shift=-4pt]
    coordinates {(N,0.697)[0.70] (PN,0.889)[0.89] (DFT,0.869)[0.87]};
\addplot[fill=vizmauve, bar shift=4pt]
    coordinates {(N,0.104)[0.10] (PN,0.828)[0.83] (DFT,0.104)[0.10]};
\draw[vizthree, line width=1pt]
    ({rel axis cs:0,0}|-{axis cs:N,0.947}) -- ({rel axis cs:1,0}|-{axis cs:N,0.947});
\node[anchor=north west, font=\tiny, text=vizthree] at (rel axis cs:0.04,1.0) {Clean $0.95$};

\nextgroupplot[title={UniTok}]
\addplot[fill=vizone, bar shift=-4pt]
    coordinates {(N,0.856)[0.86] (PN,0.902)[0.90] (DFT,0.890)[0.89]};
\addplot[fill=vizmauve, bar shift=4pt]
    coordinates {(N,0.409)[0.41] (PN,0.835)[0.84] (DFT,0.409)[0.41]};
\draw[vizthree, line width=1pt]
    ({rel axis cs:0,0}|-{axis cs:N,0.937}) -- ({rel axis cs:1,0}|-{axis cs:N,0.937});
\node[anchor=north west, font=\tiny, text=vizthree] at (rel axis cs:0.04,1.0) {Clean $0.94$};

\nextgroupplot[title={FSQ}]
\addplot[fill=vizone, bar shift=-4pt]
    coordinates {(N,0.775)[0.78] (PN,0.773)[0.77] (DFT,0.880)[0.88]};
\addplot[fill=vizmauve, bar shift=4pt]
    coordinates {(N,0.596)[0.60] (PN,0.597)[0.60] (DFT,0.596)[0.60]};
\draw[vizthree, line width=1pt]
    ({rel axis cs:0,0}|-{axis cs:N,0.831}) -- ({rel axis cs:1,0}|-{axis cs:N,0.831});
\node[anchor=north west, font=\tiny, text=vizthree] at (rel axis cs:0.04,1.0) {Clean $0.83$};

\end{groupplot}
\end{tikzpicture}
\caption{No defense protects both paths at once. Reconstruction-path (\cbox{vizone}) and probe-path (\cbox{vizmauve}) AUC under M2-unb, $\epsilon=8/255$, mean over datasets, for None (N), Pixel-noise (PN), and Decoder-FT (DFT). Green line marks each tokenizer's clean baseline (value labeled top-left of each panel). Decoder-FT recovers reconstruction but not the probe. Pixel-noise raises both, and only for MedITok and UniTok ($n{=}128$ per dataset).}
\label{fig:defense-grid}
\end{figure}
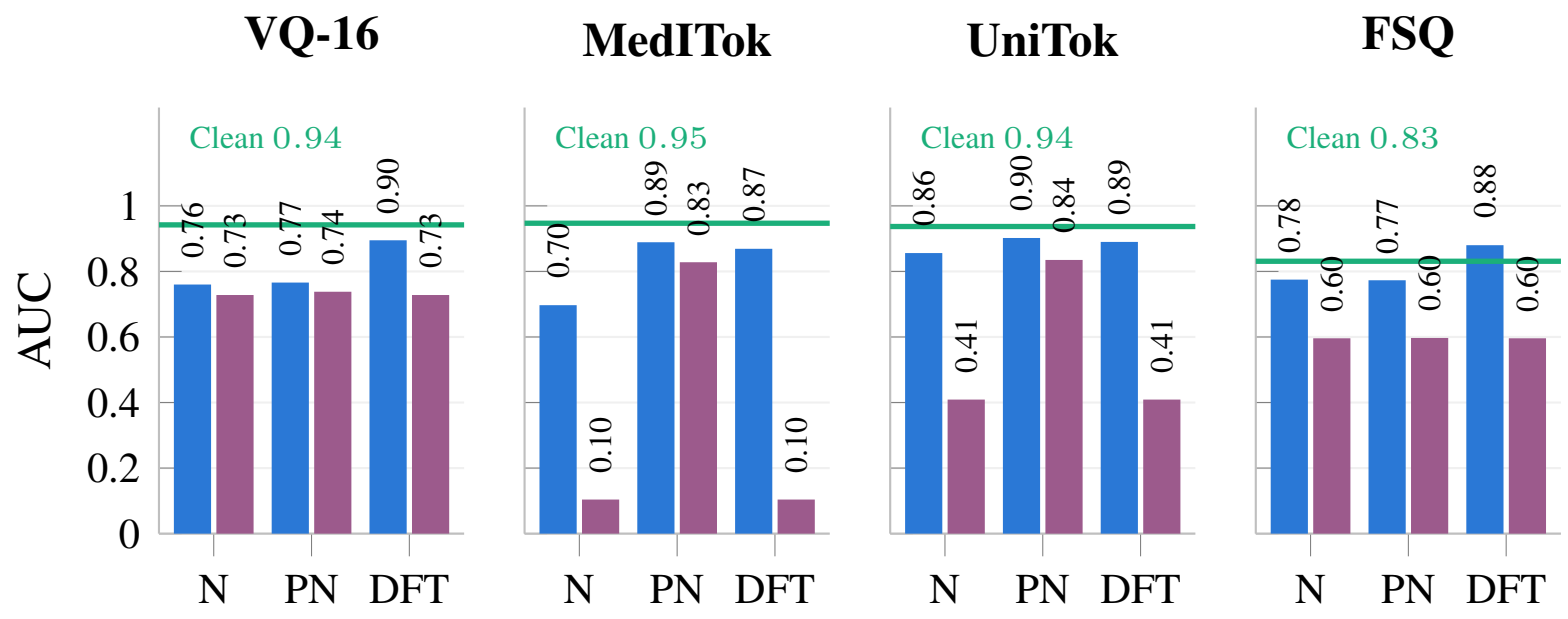

Figure~\ref{fig:defense-grid} visualizes the reconstruction and probe tradeoff across all four tokenizers for the three main defenses.

\textbf{Adaptive robustness.}
Static evaluation overstates defense strength when the adversary can compute gradients through the defense, a failure mode known as obfuscated gradients \citep{athalye2018obfuscated}. We evaluate this with two adaptive attacks:

\begin{itemize}
    \item \textbf{BPDA} (straight-through gradient estimation through quantization) collapses the reconstruction AUC of Encoder-FT, Decoder-FT, and Encoder-FT+Decoder-FT from $0.858$--$0.996$ in static evaluation to approximately $0.000$ on every tokenizer except FSQ, which retains $0.014$--$0.781$ on APTOS and ISIC. The static improvements therefore reflect gradient obfuscation rather than genuine robustness.

    \item \textbf{Encoder-FT cross-family generalization.} Encoder-FT resists an attacker using the same margin objective it was trained against (mean AUC $0.800$--$0.915$), but collapses under BPDA's decode-and-classify loss and a held-out Carlini-Wagner $\ell_2$ attack (mean AUC $\approx 0.000$ on learned-codebook tokenizers; FSQ retains $0.234$--$0.264$). Its robustness generalizes within the attack family used for training but not across attack families.
\end{itemize}

\textbf{Formal certification.}
SQS provides a certified radius guarantee \citep{cohen2019certified} rather than an empirical result a stronger attacker could overturn. Certification rate alone can nonetheless be misleading: FSQ achieves $\geq 98\%$ certification rate on Kermany and APTOS while its accuracy is exactly $0.500$, at random chance. Full per-defense numbers, the $\epsilon$ grid, and certification rates and radii are in Appendix~\ref{app:full-sweeps}.

\textbf{Reconstruction fidelity.}
Sections~\ref{sec:experiments}--\ref{sec:defenses} measure damage as classifier
AUC. Measured instead on the reconstructed image (LPIPS/PSNR/SSIM vs.\ clean;
full tables in Appendix~\ref{app:full-sweeps}), the same path-dependence
holds where fidelity degrades monotonically with budget, and M2-unb is
consistently more damaging than M2-sat at same budget (MedITok LPIPS
$0.163\!\to\!0.472$ at $\epsilon{=}8/255$) with continued displacement past the
boundary costs reconstruction quality, not just a flipped code. Decoder-FT is
the strongest fidelity defense; Encoder-FT helps fidelity but far less than Decoder-FT.

\FloatBarrier

\section{Limitations}
\label{sec:limitations}
%%%%%%%%%%%%%%%%%%%%%%%%%%%%%%%%%%%%%%%%%%%%%%%%%%%%%%%%%%%%

Results are means over three datasets with small subsets ($n{\leq}256$ per attack
cell, $n{=}128$ per defense cell), without confidence intervals or significance tests,
so small differences between cells should not be over-interpreted; our main conclusions
rest on large effects (e.g., reconstruction-path AUC $\leq 0.013$ and token-path AUC
$0.919\rightarrow0.239$). The generative evaluation has a narrow scope (one
vision-language model, LLaVA-Med with MedITok; inference-only; $20$ images/dataset),
and we red-team the generative model but test no defense on it, so whether the behavior
holds for other architectures, and whether our defenses transfer there, is unknown. We
also do not compare directly against the embedding-displacement attack of
\citet{bhagwatkar2026adversarial} under identical conditions. VQ-16's margin was computed on unnormalized
latents while its quantizer normalizes them, though all reported metrics use the true
quantizer. The adaptive suite covers only BPDA and one held-out Carlini-Wagner check.
SQS is not combined with Decoder-FT and reconstruction fidelity was measured only for
the fine-tuning defenses, not SQS or Pixel-noise.

\section{Discussion and conclusion}
\label{sec:discussion}
%%%%%%%%%%%%%%%%%%%%%%%%%%%%%%%%%%%%%%%%%%%%%%%%%%%%%%%%%%%%

Discrete image tokenizers are now upstream of many deployed medical AI systems, from
classifiers to generative vision-language models, connecting pixel-space
perturbations directly to diagnostic outcomes through a quantization step that has
been almost entirely overlooked in the adversarial robustness literature. Our
red-teaming shows this step is a genuine attack surface: an imperceptible, label-free
perturbation, computed from public tokenizer weights alone, propagates through the
tokenizer to make a deployed medical VQA model invent or erase disease.

The central finding is that robustness at this stage is not a single property. The
same perturbation causes qualitatively different damage depending on which downstream
path it travels, collapsing reconstruction while sparing the token path, or the
reverse. We term this \emph{path-dependence}. It means no single robustness number
characterizes a tokenizer, and a defense optimized for one path cannot be assumed to
generalize to the other. The mechanism of harm is displacement past the quantization
boundary, not the token flip itself, so a near-$100\%$ flip rate is uninformative
about diagnostic risk. In our comparison, the medical-domain tokenizer pairs the best clean accuracy
with the worst token-path fragility, and none of the six defenses we evaluate closes both failure
modes; several collapse once the attacker adapts.

For clinical deployment, the implications are concrete. First, a tokenizer's
robustness must be evaluated on the specific downstream path a system actually uses,
against an adaptive adversary, and reported as diagnostic outcome rather than token
statistics; a tokenizer that looks robust on one path can be destroyed on the other.
Second, because the vulnerable component is shared and its weights are public, a
single weakness can propagate to models built on it, as we show for LLaVA-Med; whether
other generative systems that a clinician might trust to read a scan, or that a patient
might query directly, inherit it remains to be tested. Third, the same imperceptible
perturbations may also arise from ordinary acquisition, compression, and transmission
artifacts, so both failure modes can occur without any adversary in deployed clinical
pipelines. As medical AI moves toward multimodal and report-generating systems built
on discrete tokenizers, characterizing and defending the quantization step is a
prerequisite for safe deployment.

\section{Broader impact}
\label{sec:broader-impact}

Red-teaming a component of deployed medical AI carries dual-use risk. We mitigate it
by attacking only public, already-released tokenizers and by pairing every attack
with a defense evaluation, so the net contribution is defensive. The risk we
emphasize is not a novel attack but false confidence: defenses that look robust yet
collapse under an adaptive adversary, and tokenizers assumed robust on one downstream
path but not another. Because the same imperceptible perturbations can also arise from
ordinary acquisition, compression, and transmission noise, these failures threaten
any system built on the tokenizer, whether a clinician relies on it in the loop or a
patient queries it directly, even absent a deliberate adversary.

\begin{ack}
This work was conducted through the Algoverse AI Research program, which also provided the compute resources. The authors received no external funding for this work.
\end{ack}

\bibliographystyle{plainnat}
\bibliography{references}

\newpage

\appendix
\begin{center}
{\Large \textbf{Appendix}}
\end{center}

\vspace{1em}

\section{Hyperparameters and experimental setup}
\label{app:hyperparams}
%%%%%%%%%%%%%%%%%%%%%%%%%%%%%%%%%%%%%%%%%%%%%%%%%%%%%%%%%%%%

\begin{comment}
Table~\ref{tab:hyperparams} lists every hyperparameter shared across tokenizers and
datasets. \textbf{Why $256$ then $128/128$, not $256$ throughout:} attack evaluation
(Section~\ref{sec:experiments}) uses the full $256$-image subset because no
hyperparameter is being chosen from it. Defense evaluation instead needs to both select
a hyperparameter (e.g.\ SQS's $\sigma$, Encoder-FT's checkpoint epoch) and report a
number for it, so the same $256$-image subset is split once, by a seeded stratified
split, into two disjoint $128$-image halves: \textsc{Sel} for fine-tuning and
hyperparameter selection, \textsc{Report} for every number in
Tables~\ref{tab:defense_recon_all}--\ref{tab:certified_radii}. A defense is never
scored on the same images used to pick it.
\end{comment}

Unless otherwise stated, all experiments in our paper use the shared configuration in Table~\ref{tab:hyperparams}. Attack evaluation is performed on a
fixed 256-image subset per dataset. For defense experiments, this subset
is split once using a seeded stratified split into disjoint 128-image
\textsc{Sel} and \textsc{Report} sets. \textsc{Sel} is used for
fine-tuning and hyperparameter selection, while all reported defense
results are computed only on \textsc{Report}.

\begin{table}[htbp]
\centering
\small
\caption{Hyperparameters and experimental setup. Shared across four tokenizers and three datasets unless otherwise noted. Code, checkpoints, extended results, and mathematical formulations are publicly available at \url{https://github.com/akashyall34/tokenizer-adversarial-robustness}.}
\label{tab:hyperparams}
\begin{tabular}{ll}
\toprule
Parameter & Value \\
\midrule

\multicolumn{2}{l}{\textbf{General setup}} \\
GPU & 1$\times$ NVIDIA A100-SXM4-80GB \\
Software & PyTorch $\geq$2.1, torchvision $\geq$0.16, timm $1.0.8$ \\
Image size & $256\times256$ \\
Random seed & $42$ \\
Attack evaluation & $256$ images / dataset \\
Defense selection / report & $128$ / $128$ images \\

\midrule
\multicolumn{2}{l}{\textbf{PGD attacks}} \\
$\ell_\infty$ budget $\epsilon$ & $\{0.5,1,2,4,8\}/255$ \\
Steps / learning rate & $50$ / $0.01$ \\
Batch size & $4$--$8$ \\

\midrule
\multicolumn{2}{l}{\textbf{Encoder-FT}} \\
Training images & $128$ \\
Learning rate / batch size & $10^{-5}$ / $32$ \\
Max epochs & $40$ \\
Early stopping & patience $2$, minimum $3$ epochs \\
Improvement threshold & $10\%$ relative \\
Clean-AUC guard & maximum $10\%$ drop \\

\midrule
\multicolumn{2}{l}{\textbf{Decoder-FT}} \\
Training images & $128$ \\
Training budget & $8/255$ \\
Decoder learning rate &
$10^{-4}$ (VQ-16, FSQ); $10^{-5}$ (MedITok, UniTok) \\
Discriminator learning rate & $10^{-4}$ \\
Batch size / max epochs & $8$ / $40$ \\
Loss weights (adv., perc., recon., robust) &
$0.1,\ 1.0,\ 1.0,\ 1.0$ \\

\midrule
\multicolumn{2}{l}{\textbf{SQS}} \\
Noise scale $\sigma$ &
$\{0.0005,0.001,0.002,0.005,0.01,0.02\}$ \\
Additional flat-sweep $\sigma$ & $\{0.05,0.1\}$ \\
Vote count $K$ & $\{4,8,16,32\}$ \\
Spatial weight & $\{0.4,0.5,0.6,0.7,0.8\}$ \\
Agreement threshold (\%) & $\{50,60,70,80,90\}$ \\
Adaptive multiplier & $\{2,4,8\}\times\sigma_{\mathrm{base}}$ \\
Logit-ensemble $\sigma$ & $\{0.005,0.01,0.02\}$ \\

\midrule
\multicolumn{2}{l}{\textbf{Adaptive attacks and certification}} \\
BPDA steps / learning rate & $50$ / $0.01$ \\
CW-$\ell_2$ budget & $\{4,8,16\}$ \\
CW confidence $\kappa$ & $0$ \\
CW steps / learning rate & $50$ / $0.01$ \\
Certification draws $N$ & $100$ \\
Clopper--Pearson $\alpha$ & $0.001$ \\

\bottomrule
\end{tabular}

\end{table}

\begin{comment}
Encoder-FT and Decoder-FT both fine-tune against cached adversarial images already
produced by the attack sweep (Section~\ref{sec:defenses}), not against a fresh attack
generated during training; Encoder-FT trains one checkpoint per $\epsilon$ in the
budget grid, while Decoder-FT trains a single checkpoint at the worst-case budget.
Both use the early-stop rule above, applied to their own training loss, and both use
the \textsc{Sel} half of the split described above, never the \textsc{Report} half.
Attack batch size and Decoder-FT's decoder learning rate are the two settings that are
not shared across all four tokenizers: batch size drops from $8$ to $4$ for APTOS and
for MedITok/UniTok on ISIC, and Decoder-FT's decoder learning rate is $10\times$ lower
for MedITok/UniTok than for VQ-16/FSQ.
\end{comment}

Encoder-FT and Decoder-FT are trained on cached adversarial examples from
the attack sweep in Section~\ref{sec:defenses}. Encoder-FT trains one
checkpoint per attack budget, whereas Decoder-FT uses a single checkpoint
trained at $8/255$. Both use only the \textsc{Sel} split and apply early
stopping to their training loss; the \textsc{Report} split is never used
for model or hyperparameter selection. Attack batch size is reduced from
$8$ to $4$ for APTOS and for MedITok/UniTok on ISIC because of memory
requirements.

\FloatBarrier

\section{Models and datasets}
\label{app:models-datasets}
%%%%%%%%%%%%%%%%%%%%%%%%%%%%%%%%%%%%%%%%%%%%%%%%%%%%%%%%%%%%

We evaluate four tokenizers spanning the main quantization strategies (Table~\ref{tab:tokenizers}): LlamaGen VQ-16's
\citep{sun2024autoregressive} single learned codebook, representative of
the VQ-VAE family; FSQ's \citep{mentzer2024finite} codebook-free fixed
scalar rounding; and UniTok's \citep{ma2025unitok} eight parallel
cosine-similarity sub-quantizers, shared exactly by its medical-domain
counterpart MedITok \citep{ma2025meditok}. UniTok and MedITok share an architecture and differ
in pretraining data (general-domain vs.\ clinical images), giving a matched-architecture
comparison of domain-specific pretraining; other training differences may also
contribute, so it is not a fully causal test.

\begin{table}[htbp]
\centering
\small
\caption{The four tokenizers evaluated, spanning three quantization strategies. UniTok and MedITok share the same multi-codebook architecture but differ in pretraining data, giving a matched-architecture comparison of domain-specific pretraining.}
\label{tab:tokenizers}
\begin{tabular}{llll}
\toprule
Tokenizer & Quantization type & Domain & Source \\
\midrule
LlamaGen VQ-16 & Vector quantization, learned codebook & General & \citep{sun2024autoregressive} \\
FSQ & Scalar quantization, fixed thresholds & General & \citep{mentzer2024finite} \\
UniTok & Multi-codebook vector quantization & General & \citep{ma2025unitok} \\
MedITok & Same as UniTok & Medical & \citep{ma2025meditok} \\
\bottomrule
\end{tabular}

\end{table}

Our evaluation spans three medical imaging datasets (Table~\ref{tab:datasets}),
chosen to test whether our findings transfer across imaging modalities, classifier architectures, and
diagnostic tasks; in every case the underlying multi-class labels are
reduced to one clinically meaningful binary decision. Radiography, fundus
photography, and dermoscopy differ in acquisition modality, image
statistics, and class structure, so agreement across all three is evidence
that a tokenizer vulnerability is a property of discrete quantization
itself, not of any single domain. Full dataset details, including collection,
grading criteria, and splits, are in Appendix~\ref{app:datasets}.

\begin{table}[htbp]
\centering
\small
\setlength{\tabcolsep}{3pt}
\caption{Three medical imaging datasets spanning different modalities, diagnostic tasks, and classifier architectures. Clean AUC is measured on the unperturbed test subset used throughout all experiments.}
\label{tab:datasets}
\begin{tabular}{lllcc}
\toprule
Dataset (modality) & Task & Classifier & AUC & Source \\
\midrule
Kermany (X-ray) & Pneumonia vs.\ normal & DenseNet-121 & 0.954 & \citep{kermany2018identifying} \\
APTOS 2019 (fundus) & Referable DR vs.\ not & EfficientNet-B5 & 0.993 & \citep{aptos2019} \\
ISIC 2019 (derm.) & Malignant vs.\ benign & EfficientNet-B4 & 0.922 & \citep{isic2019,tschandl2018ham10000} \\
\bottomrule
\end{tabular}
\vspace{6pt}

\end{table}

Each dataset is scored along both inference paths (reconstruction
and probe) on a fixed, class-balanced
$256$-image subset (seed $42$) held constant across every attack. Defense
evaluation (Section~\ref{sec:defenses}) further splits this into seeded
$128$-image halves, so no defense is scored on the images used
to tune it. Attack budgets, probe dimensionalities, and full hyperparameters
are in Appendix~\ref{app:hyperparams}.

\FloatBarrier

\section{Metrics}
\label{app:metrics-detail}
%%%%%%%%%%%%%%%%%%%%%%%%%%%%%%%%%%%%%%%%%%%%%%%%%%%%%%%%%%%%

We track three quantities per image, averaged over the evaluation set
(Table~\ref{tab:metrics}). A fourth metric, certified radius, applies only to SQS and is defined where it is used (Section~\ref{sec:defenses}). Section~\ref{sec:defenses}'s reconstruction-fidelity paragraph additionally reports LPIPS, PSNR, and SSIM between the attacked reconstruction and the clean original. FID is not reported (Appendix~\ref{app:full-sweeps}) because $n{=}256$ is below the sample size FID needs to be reliable.

\begin{table}[htbp]
\centering
\small
\caption{Three evaluation metrics used throughout. Token flip rate and embedding displacement are measured at the quantizer boundary; downstream AUC is measured at the classifier output along each inference path independently.}
\label{tab:metrics}
\begin{tabular}{lp{9.2cm}}
\toprule
Metric & Measures \\
\midrule
Token flip rate & Share of tokens reassigned to a different code after the attack \\
Embedding displacement & Distance the pre-quantization representation moves in latent space; keeps growing after a flip, so values are only comparable within one tokenizer \\
Downstream diagnostic AUC & Classifier AUC on attacked vs.\ clean images, on the same test set; chosen over accuracy for being threshold-free and robust to class imbalance \\
\bottomrule
\end{tabular}

\end{table}

\FloatBarrier

\section{Dataset details}
\label{app:datasets}

\begin{table}[htbp]
\centering
\small
\setlength{\tabcolsep}{3pt}
\caption{Dataset metadata. All attack evaluations use a class-balanced subset of $256$ images ($128$ per class). The ISIC split is generated once with a fixed random seed.}
\label{tab:datasets-appendix}
\begin{tabular}{>{\raggedright\arraybackslash}p{4.0cm}l>{\raggedright\arraybackslash}p{2.9cm}rrrr}
\toprule
Dataset & Modality & Task & Total & Train & Val & Test \\
\midrule
Kermany \citep{kermany2018identifying,kermany2018labeled}
  & Chest X-ray & Pneumonia vs.\ Normal
  & $5{,}856$ & $4{,}709$ & $523$ & $624$ \\
APTOS 2019 \citep{aptos2019}
  & Retinal fundus & Referable DR (grades 2--4)
  & $3{,}662$ & $2{,}930$ & $366$ & $366$ \\
ISIC 2019 \citep{isic2019,tschandl2018ham10000,codella2017isic,combalia2019bcn20000}
  & Dermoscopy & Malignant vs.\ Benign
  & $25{,}331$ & $17{,}732$ & $2{,}533$ & $5{,}066$ \\
\bottomrule
\end{tabular}
\vspace{4pt}

\end{table}

\FloatBarrier

\section{Generative downstream: full results}
\label{app:vqa}
%%%%%%%%%%%%%%%%%%%%%%%%%%%%%%%%%%%%%%%%%%%%%%%%%%%%%%%%%%%%

\paragraph{Setup.} We attack the MedITok vision tower of a frozen LLaVA-Med with
the objectives of Section~\ref{sec:attacks} unchanged (Adam, $50$ steps,
$\mathrm{lr}=10^{-2}$, $\ell_\infty$ projection), then pass the perturbed image to
the full model and read its answer. Nothing is fine-tuned, and the attack never
observes the question, the answer, or any gradient from the language decoder.
We use $20$ images per dataset, balanced across classes, three questions each, at
$\epsilon\in\{2,4,8\}/255$. M1 is not applicable here: it attacks through a
classification head, and this path has no probe.

\paragraph{Counting an answer correct.} An answer is correct when it agrees with
the dataset label: for a diagnostic question its stated verdict must match the
ground-truth class, and for a modality or body-region question the modality or
anatomy it names must match the dataset's. We read the verdict the answer declares
rather than searching for keywords, so an answer naming a different modality is not
counted correct for mentioning the right word in passing.

\subsection*{The attack needs budget, not flips}

Table~\ref{tab:vqa-eps} is the central result. At the smallest budget the tokens
are already almost entirely flipped, yet most answers survive. As the budget grows
the flip rate barely moves while the answers collapse. What does grow with the
budget is displacement, and M2-sat makes the same point from the other side: it
flips just as many tokens at every budget, displaces about half as far, and leaves
the answers alone.

\begin{table}[htbp]
\centering
\small
\setlength{\tabcolsep}{6pt}
\caption{Flip rate, displacement and answer accuracy on the modality and
body-region questions, pooled over the three datasets ($n{=}99$ image-question
cells the model answers correctly when clean). Flip rate is saturated from
$\epsilon{=}2$ and cannot explain the collapse; displacement tracks it.}
\label{tab:vqa-eps}
\begin{tabular}{lcccccccc}
\toprule
& \multicolumn{3}{c}{M2-unb} & \multicolumn{3}{c}{M2-sat} & \multicolumn{2}{c}{Random} \\
\cmidrule(lr){2-4}\cmidrule(lr){5-7}\cmidrule(lr){8-9}
$\epsilon$ & Flip\% & Disp. & Correct & Flip\% & Disp. & Correct & Flip\% & Correct \\
\midrule
$2/255$ &  98.7 & 50.6 & $59.0\%$ &  99.5 & 24.9 & $96.0\%$ & 28.3 & $99.0\%$ \\
$4/255$ &  99.8 & 65.1 & $13.0\%$ &  99.9 & 32.2 & $97.0\%$ & 54.5 & $99.0\%$ \\
$8/255$ & 100.0 & 71.3 & $\phantom{0}1.0\%$ & 100.0 & 38.6 & $98.0\%$ & 74.6 & $98.0\%$ \\
\bottomrule
\end{tabular}
\end{table}

\begin{table}[htbp]
\centering
\footnotesize
\setlength{\tabcolsep}{4pt}
\renewcommand{\arraystretch}{1.1}
\caption{The same effect on single images. Each answer is verbatim under M2-unb.
At $\epsilon{=}2/255$ the token sequence is already $98\%$ different from the
clean one and the answer is still \textcolor{vizthree}{right}; it only turns
\textcolor{vizred}{wrong} once displacement grows.}
\label{tab:vqa-progression}
\begin{tabular}{@{}p{2.4cm} l p{7.9cm}@{}}
\toprule
Image & $\epsilon$ (flip\%, disp.) & Answer under M2-unb \\
\midrule
\multirow{3}{=}{Chest X-ray \newline \emph{Which part of the body?}}
 & $2$ \ ($98.0$, $43.4$) & \textcolor{vizthree}{This image belongs to the \textbf{thoracic region} of the body, specifically showing the thoracic aorta and its branches\ldots} \\
 & $4$ \ ($100.0$, $57.9$) & \textcolor{vizred}{This image belongs to the \textbf{abdominal region} of the body. The presence of the liver, spleen\ldots} \\
 & $8$ \ ($100.0$, $66.7$) & \textcolor{vizred}{This image belongs to the \textbf{abdominal region}, specifically showing the liver and surrounding structures.} \\
\midrule
\multirow{3}{=}{Fundus \newline \emph{Which part of the body?}}
 & $2$ \ ($98.0$, $46.0$) & \textcolor{vizthree}{This image belongs to the \textbf{eye}, specifically showing the retina and optic disc.} \\
 & $4$ \ ($100.0$, $63.4$) & \textcolor{vizred}{This image belongs to the \textbf{brain}. The brain is a vital organ located within the skull\ldots} \\
 & $8$ \ ($100.0$, $69.0$) & \textcolor{vizred}{This image belongs to the \textbf{abdominal region}, specifically showing the liver and surrounding structures.} \\
\midrule
\multirow{3}{=}{Dermoscopy \newline \emph{What modality?}}
 & $2$ \ ($98.0$, $42.3$) & \textcolor{vizthree}{The image is a \textbf{dermoscopic} image, which is a specialized imaging technique that uses a dermoscope\ldots} \\
 & $4$ \ ($100.0$, $67.6$) & \textcolor{vizred}{The image is a \textbf{computed tomography (CT)} scan of the abdomen.} \\
 & $8$ \ ($100.0$, $78.9$) & \textcolor{vizred}{The image is a \textbf{histopathological slide}, a microscopic examination of stained tissue.} \\
\bottomrule
\end{tabular}
\end{table}

\subsection*{Per-question and per-direction detail}

\begin{table}[htbp]
\centering
\small
\setlength{\tabcolsep}{4pt}
\caption{Answer accuracy (\%) by dataset and question, $n{=}20$ images per row.}
\label{tab:vqa-full}
\begin{tabular}{llccccccccc}
\toprule
& & & \multicolumn{3}{c}{M2-unb} & \multicolumn{3}{c}{M2-sat} & \multicolumn{2}{c}{Random} \\
\cmidrule(lr){4-6}\cmidrule(lr){7-9}\cmidrule(lr){10-11}
Dataset & Question & Clean & $2$ & $4$ & $8$ & $2$ & $4$ & $8$ & $2$ & $8$ \\
\midrule
\multirow{3}{*}{Kermany} & Modality & 100.0 & 75.0 &  5.0 & 0.0 & 100.0 & 100.0 & 100.0 & 100.0 & 100.0 \\
 & Body region             & 100.0 & 90.0 & 60.0 & 5.0 & 100.0 & 100.0 & 100.0 & 100.0 & 100.0 \\
 & Pneumonia$^{\dagger}$   &  50.0 & 50.0 & 50.0 & 30.0 & 55.0 & 50.0 & 47.4 & 50.0 & 50.0 \\
\midrule
\multirow{3}{*}{APTOS} & Modality   & 100.0 & 55.0 &  0.0 & 0.0 & 100.0 & 100.0 & 100.0 & 100.0 & 100.0 \\
 & Body region             & 100.0 & 65.0 &  0.0 & 0.0 & 100.0 & 100.0 & 100.0 & 100.0 & 100.0 \\
 & Referable DR            &  85.0 & 45.0 & 25.0 & 40.0 & 80.0 & 85.0 & 70.0 & 85.0 & 90.0 \\
\midrule
\multirow{2}{*}{ISIC} & Modality    &  95.0 & 10.0 &  0.0 & 0.0 & 80.0 & 85.0 & 90.0 & 95.0 & 90.0 \\
 & Benign/malignant        &  80.0 & 45.0 & 50.0 & 40.0 & 80.0 & 70.0 & 60.0 & 75.0 & 80.0 \\
\bottomrule
\end{tabular}

\vspace{3pt}
{\footnotesize $^{\dagger}$The model answers ``no'' on all $20$ chest X-rays,
including the $10$ with pneumonia, so its $50\%$ is the score of always answering
no. There is no correct diagnosis for the attack to destroy, and we exclude the
question from the main text. The ISIC lesion-type question is excluded for the same
reason: on benign lesions its clean answers include melanoma, basal cell carcinoma
and verrucous carcinoma.}
\end{table}

\begin{table}[htbp]
\centering
\small
\setlength{\tabcolsep}{6pt}
\caption{Direction of diagnostic error under M2-unb, on images answered correctly
when clean. The attack never sees the label and moves the answer either way; which
direction dominates differs by dataset. These are counts, not incidence rates.}
\label{tab:vqa-direction}
\begin{tabular}{llccc}
\toprule
Dataset & Error direction & $\epsilon{=}2$ & $4$ & $8$ \\
\midrule
\multirow{2}{*}{APTOS}
 & Disease erased (false negative)   & 4/10 & 6/10 & 3/10 \\
 & Disease invented (false positive) & 6/7  & 7/7  & 7/7  \\
\midrule
\multirow{2}{*}{ISIC}
 & Malignant read as benign & 6/6  & 6/6  & 6/6  \\
 & Benign read as malignant & 1/10 & 0/10 & 2/10 \\
\bottomrule
\end{tabular}
\end{table}

\paragraph{Scope.} These results cover MedITok only, because it is the only
tokenizer in our study with a released multimodal checkpoint, and are
inference-only. The diagnostic questions have few clean-correct images ($17$ for
APTOS, $16$ for ISIC), so Table~\ref{tab:vqa-direction} reports counts rather than
rates. Greedy decoding is sensitive to small numerical differences in the encoder,
so wording varies between runs and the smallest breaking $\epsilon$ for an
individual image can shift by one step; the pooled rates are stable, and the quotes
in Figure~\ref{fig:vqa} come from the run that produced the displayed images.

%%%%%%%%%%%%%%%%%%%%%%%%%%%%%%%%%%%%%%%%%%%%%%%%%%%%%%%%%%%%
\section{Extended results and analysis}
\label{app:empirical}
%%%%%%%%%%%%%%%%%%%%%%%%%%%%%%%%%%%%%%%%%%%%%%%%%%%%%%%%%%%%

\subsection{Flips versus damage}
\label{app:flip-saturation}

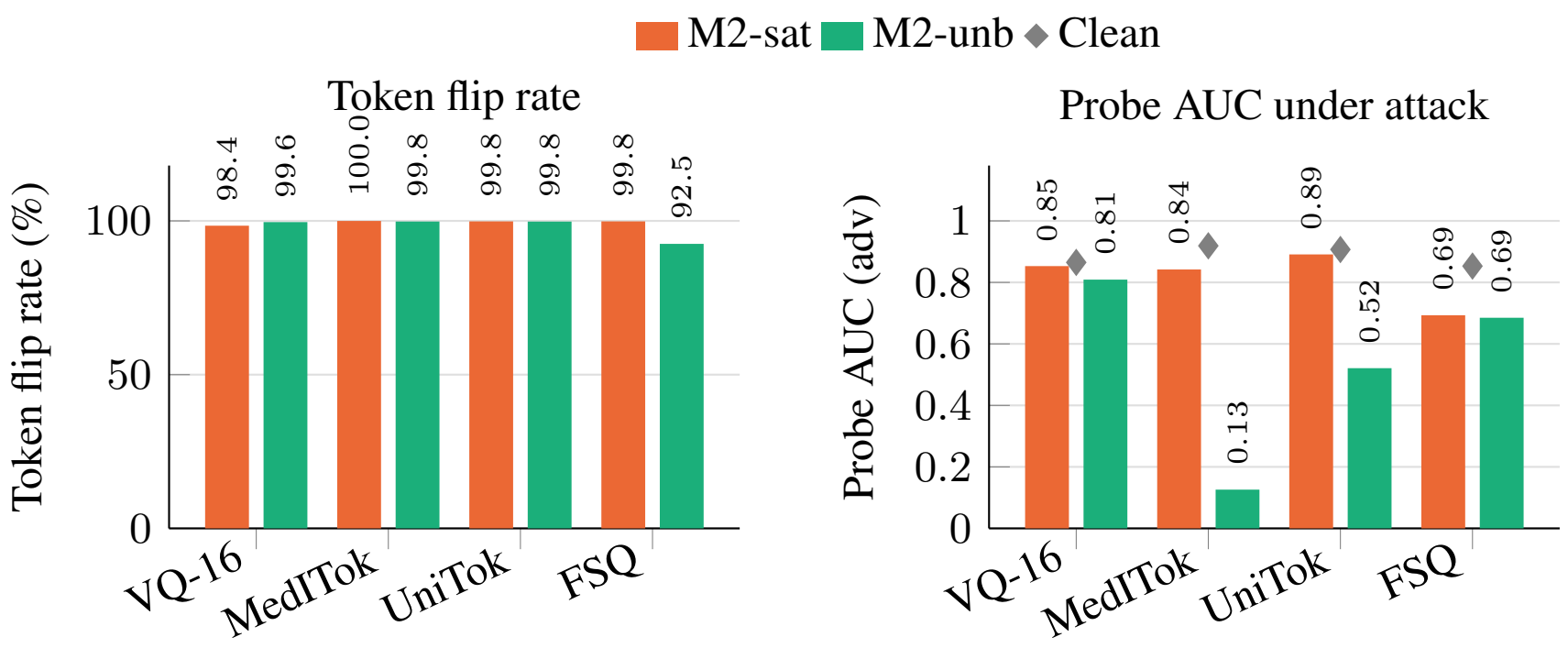
\begin{figure}[htbp]
\centering
\begin{tikzpicture}
\begin{groupplot}[
    group style={group size=2 by 1, horizontal sep=1.8cm},
    width=5.7cm, height=4.2cm,
    symbolic x coords={VQ-16,MedITok,UniTok,FSQ},
    xtick=data,
    x tick label style={rotate=25, anchor=east, font=\small},
    enlarge x limits=0.22,
    axis lines*=left,
    ymajorgrids, grid style={gray!25},
    tick label style={font=\small},
    label style={font=\small},
    title style={font=\small},
    ]
\nextgroupplot[title={Token flip rate}, ylabel={Token flip rate (\%)}, ymin=0, ymax=118, bar width=9pt, ybar]
\addplot[fill=viztwo, draw=none, barlegend=viztwo, bar shift=-6pt,
    nodes near coords, nodes near coords style={font=\tiny, rotate=90, anchor=west, xshift=1.5pt,
        /pgf/number format/fixed, /pgf/number format/precision=1, /pgf/number format/zerofill}]
    coordinates {(VQ-16,98.44) (MedITok,99.98) (UniTok,99.84) (FSQ,99.84)};
\addplot[fill=vizthree, draw=none, barlegend=vizthree, bar shift=6pt,
    nodes near coords, nodes near coords style={font=\tiny, rotate=90, anchor=west, xshift=1.5pt,
        /pgf/number format/fixed, /pgf/number format/precision=1, /pgf/number format/zerofill}]
    coordinates {(VQ-16,99.61) (MedITok,99.80) (UniTok,99.78) (FSQ,92.53)};
\nextgroupplot[title={Probe AUC under attack}, ylabel={Probe AUC (adv)}, ymin=0, ymax=1.18, ytick={0,0.2,0.4,0.6,0.8,1}, bar width=9pt, ybar,
    legend style={at={(-0.15,1.25)},anchor=south,legend columns=-1,draw=none,font=\small}]
\addplot[fill=viztwo, draw=none, barlegend=viztwo, bar shift=-6pt,
    nodes near coords, nodes near coords style={font=\tiny, rotate=90, anchor=west, xshift=1.5pt,
        /pgf/number format/fixed, /pgf/number format/precision=2, /pgf/number format/zerofill}]
    coordinates {(VQ-16,0.853) (MedITok,0.842) (UniTok,0.891) (FSQ,0.693)};
\addplot[fill=vizthree, draw=none, barlegend=vizthree, bar shift=6pt,
    nodes near coords, nodes near coords style={font=\tiny, rotate=90, anchor=west, xshift=1.5pt,
        /pgf/number format/fixed, /pgf/number format/precision=2, /pgf/number format/zerofill}]
    coordinates {(VQ-16,0.809) (MedITok,0.126) (UniTok,0.521) (FSQ,0.685)};
\addplot[only marks, mark=diamond*, mark size=2.5pt, color=gray,
    legend image code/.code={\node[diamond, fill=gray, scale=0.55] at (0.3cm,0cm) {};}]
    coordinates {(VQ-16,0.865) (MedITok,0.919) (UniTok,0.907) (FSQ,0.853)};
\legend{M2-sat,M2-unb,Clean}
\end{groupplot}
\end{tikzpicture}
\caption{Flips do not imply damage. Bounded (M2-sat) and
unbounded (M2-unb) geometric attacks at $\epsilon=4/255$, mean over
datasets. Left: token flip rates are near-identical. Right: adversarial
probe AUC diverges sharply; gray diamonds mark each tokenizer's clean
(unattacked) probe AUC. Same flips, opposite diagnostic outcomes.}
\label{fig:flips}
\end{figure}

Figure~\ref{fig:flips} expands the mechanism summarized by the flip column
of Table~\ref{tab:main}: the bounded and unbounded geometric attacks reach
near-identical token flip rates, yet only the unbounded variant, which
keeps displacing the representation past the boundary, collapses the probe.

\FloatBarrier

\subsection{Cross-tokenizer generalization across modalities}
\label{app:cross-tokenizer}

\begin{figure}[htbp]
\centering
\begin{tikzpicture}
\begin{groupplot}[
    group style={group size=2 by 1, horizontal sep=2.0cm},
    width=6.2cm, height=4.8cm,
    xmode=log, log basis x=2,
    xtick={0.5,1,2,4,8}, xticklabels={0.5,1,2,4,8},
    ymin=0, ymax=1.05, ytick={0,0.2,0.4,0.6,0.8,1},
    axis lines*=left,
    ymajorgrids, grid style={gray!25},
    tick label style={font=\normalsize},
    label style={font=\normalsize},
    title style={font=\normalsize},
    xlabel={$\ell_\infty$ budget $\epsilon$ (/255)},
    ylabel={AUC},
    legend style={at={(1.05,-0.30)},anchor=north,legend columns=-1,draw=none,font=\normalsize},
    ]
\nextgroupplot[title={Token (probe) path}]
\addplot[vizone, line width=1.5pt, mark=*, mark size=1.8pt] coordinates {(0.5,0.869)(1,0.869)(2,0.836)(4,0.809)(8,0.709)};
\addplot[viztwo, line width=1.5pt, mark=square*, mark size=1.8pt] coordinates {(0.5,0.827)(1,0.601)(2,0.239)(4,0.126)(8,0.097)};
\addplot[vizthree, line width=1.5pt, mark=triangle*, mark size=2.2pt] coordinates {(0.5,0.869)(1,0.830)(2,0.721)(4,0.521)(8,0.431)};
\addplot[vizmauve, line width=1.5pt, mark=diamond*, mark size=2.2pt] coordinates {(0.5,0.844)(1,0.826)(2,0.769)(4,0.688)(8,0.653)};
\draw[vizone, dashed, line width=1.5pt] (axis cs:0.5,0.871) -- (axis cs:8,0.871);
\draw[viztwo, dashed, line width=1.5pt] (axis cs:0.5,0.913) -- (axis cs:8,0.913);
\draw[vizthree, dashed, line width=1.5pt] (axis cs:0.5,0.903) -- (axis cs:8,0.903);
\draw[vizmauve, dashed, line width=1.5pt] (axis cs:0.5,0.853) -- (axis cs:8,0.853);
\legend{VQ-16, MedITok, UniTok, FSQ}
\nextgroupplot[title={Reconstruction path}]
\addplot[vizone, line width=1.5pt, mark=*, mark size=1.8pt] coordinates {(0.5,0.942)(1,0.937)(2,0.901)(4,0.832)(8,0.752)};
\addplot[viztwo, line width=1.5pt, mark=square*, mark size=1.8pt] coordinates {(0.5,0.950)(1,0.943)(2,0.923)(4,0.804)(8,0.705)};
\addplot[vizthree, line width=1.5pt, mark=triangle*, mark size=2.2pt] coordinates {(0.5,0.947)(1,0.936)(2,0.921)(4,0.878)(8,0.849)};
\addplot[vizmauve, line width=1.5pt, mark=diamond*, mark size=2.2pt] coordinates {(0.5,0.846)(1,0.850)(2,0.850)(4,0.833)(8,0.785)};
\draw[vizone, dashed, line width=1.5pt] (axis cs:0.5,0.942) -- (axis cs:8,0.942);
\draw[viztwo, dashed, line width=1.5pt] (axis cs:0.5,0.947) -- (axis cs:8,0.947);
\draw[vizthree, dashed, line width=1.5pt] (axis cs:0.5,0.937) -- (axis cs:8,0.937);
\draw[vizmauve, dashed, line width=1.5pt] (axis cs:0.5,0.847) -- (axis cs:8,0.847);
\end{groupplot}
\end{tikzpicture}
\caption{Budget sweep under M2-unb, mean over the three datasets.
Left: token (probe) path. Right: reconstruction path. Same $\epsilon\in\{0.5,1,2,4,8\}/255$
grid used throughout this paper; dashed horizontal lines mark each tokenizer's own
clean (unattacked) AUC, same color as its solid adversarial-AUC curve; see
Tables~\ref{tab:probe_m3unb_all}--\ref{tab:recon_m3unb_all} for the
per-dataset numbers this figure averages.}
\label{fig:sweep}
\end{figure}
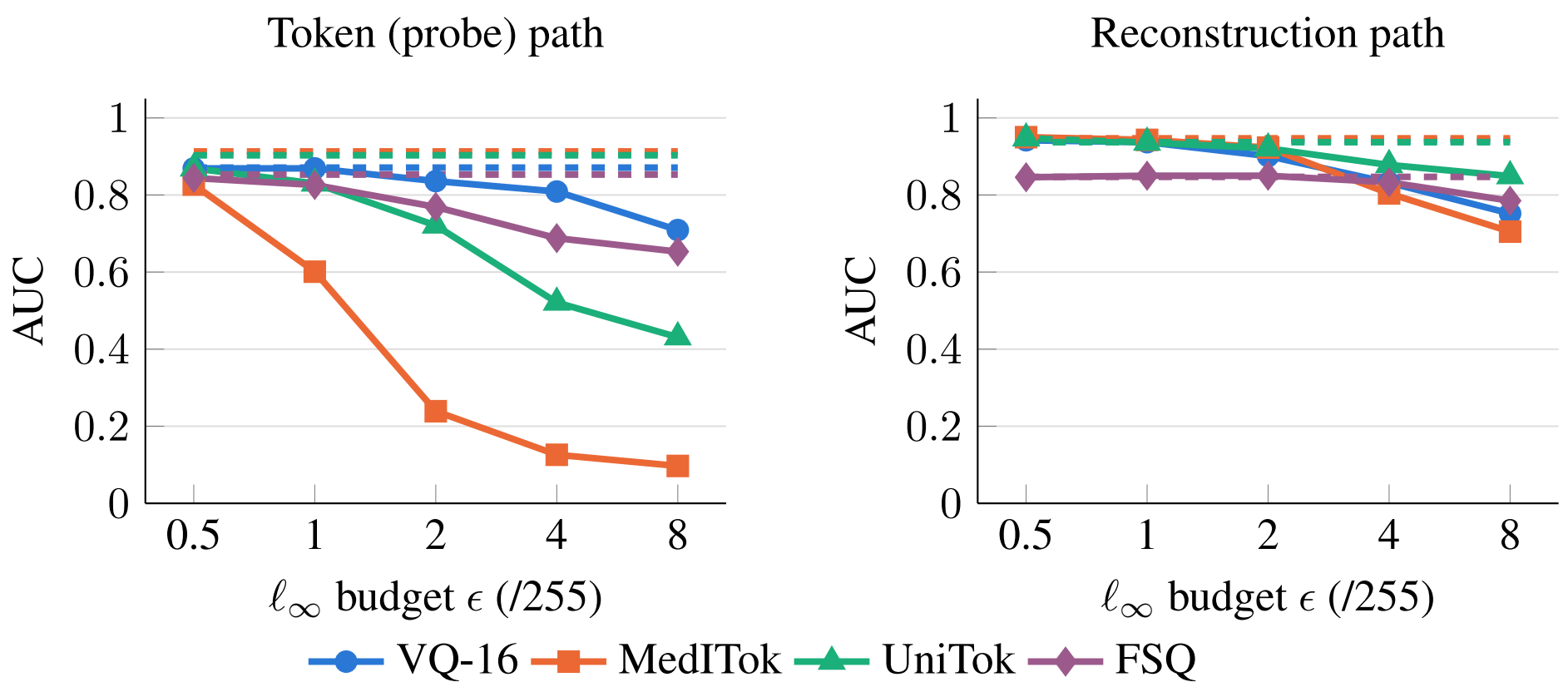

Figure~\ref{fig:sweep} sweeps the M2-unb budget on both paths. MedITok collapses fastest on the token path; all tokenizers resist on the reconstruction path until the largest budget. Tables~\ref{tab:probe_m3unb_all}--\ref{tab:recon_m3unb_all} give the per-dataset breakdown.

Table~\ref{tab:flip_vs_damage} summarizes the flip-vs-displacement dissociation at $\epsilon=2/255$: bounded and unbounded attacks reach nearly identical flip rates, yet only M2-unb's larger embedding displacement collapses the probe. Table~\ref{tab:displacement_all} reports the full displacement grid per dataset; on every learned-codebook tokenizer M2-unb displaces roughly twice as far as M2-sat at matched flip rates, stable across modalities. FSQ is the exception.

\begin{table}[htbp]
\centering
\small
\setlength{\tabcolsep}{4.5pt}
\caption{Displacement past the boundary, not the token flip, causes the damage. Mean over the three datasets at $\epsilon=2/255$. The bounded (M2-sat) and unbounded (M2-unb) geometric attacks reach comparable near-saturated flip rates, but M2-unb moves the pre-quantization embedding roughly twice as far (Emb.\ displ.), and only M2-unb collapses the probe AUC. The flip is matched; the displacement and the harm are not. Displacement is read within a tokenizer only, never across (its scale differs by an order of magnitude). $n{=}256$ per cell.}
\label{tab:flip_vs_damage}
\begin{tabular}{lccccccc}
\toprule
\multirow{2}{*}{Tokenizer} & \multirow{2}{*}{Clean AUC$_p$} & \multicolumn{2}{c}{Flip rate (\%)} & \multicolumn{2}{c}{Emb.\ displ.} & \multicolumn{2}{c}{Probe AUC (adv)} \\
\cmidrule(lr){3-4}\cmidrule(lr){5-6}\cmidrule(lr){7-8}
 & & M2-sat & M2-unb & M2-sat & M2-unb & M2-sat & M2-unb \\
\midrule
LlamaGen VQ-16 & 0.863 & 96.73 & 97.99 & 35.88 & 75.31 & 0.864 & 0.836 \\
MedITok & 0.919 & 99.29 & 98.58 & 21.61 & 50.24 & 0.868 & 0.239 \\
UniTok & 0.907 & 99.04 & 99.03 & 11.34 & 22.82 & 0.886 & 0.721 \\
FSQ & 0.855 & 96.50 & 80.62 & 1.66 & 1.82 & 0.764 & 0.769 \\
\bottomrule
\end{tabular}

\end{table}

\begin{table}[htbp]
\centering
\small
\setlength{\tabcolsep}{3.5pt}
\caption{Embedding displacement (pre-quantization), M2-sat vs.\ M2-unb, all datasets, full $\epsilon$ grid. The unbounded attack moves the representation consistently farther than the bounded one on every learned-codebook tokenizer, while flip rates are matched (Table~\ref{tab:flip_vs_damage}). FSQ is the exception: bounded and unbounded displace almost equally, consistent with its degraded baseline and lack of a token-path dissociation. Displacement is comparable within a column, not across tokenizers. $n{=}256$ per dataset.}
\label{tab:displacement_all}
\resizebox{\linewidth}{!}{%
\begin{tabular}{llcccccccccc}
\toprule
& \multirow{2}{*}{Tokenizer} & \multicolumn{2}{c}{$\nicefrac{0.5}{255}$} & \multicolumn{2}{c}{$\nicefrac{1}{255}$} & \multicolumn{2}{c}{$\nicefrac{2}{255}$} & \multicolumn{2}{c}{$\nicefrac{4}{255}$} & \multicolumn{2}{c}{$\nicefrac{8}{255}$} \\
\cmidrule(lr){3-4}\cmidrule(lr){5-6}\cmidrule(lr){7-8}\cmidrule(lr){9-10}\cmidrule(lr){11-12}
& & M2-sat & M2-unb & M2-sat & M2-unb & M2-sat & M2-unb & M2-sat & M2-unb & M2-sat & M2-unb \\
\midrule
\multirow{4}{*}{Kermany}
 & LlamaGen VQ-16 & 12.21 & 20.84 & 20.33 & 37.08 & 30.35 & 62.47 & 40.62 & 93.83 & 49.96 & 120.39 \\
 & MedITok        & 7.28 & 20.38 & 13.67 & 35.98 & 22.11 & 52.03 & 30.39 & 65.21 & 37.21 & 73.01 \\
 & UniTok         & 5.30 & 10.03 & 7.41 & 15.47 & 9.56 & 20.51 & 11.48 & 24.81 & 13.61 & 27.82 \\
 & FSQ            & 0.30 & 0.31 & 0.57 & 0.61 & 1.12 & 1.23 & 2.08 & 2.32 & 3.12 & 4.07 \\
\midrule
\multirow{4}{*}{APTOS}
 & LlamaGen VQ-16 & 16.04 & 26.88 & 25.83 & 50.14 & 37.99 & 80.21 & 47.54 & 105.47 & 56.00 & 122.21 \\
 & MedITok        & 7.91 & 15.08 & 15.15 & 28.34 & 23.09 & 45.76 & 30.40 & 60.32 & 37.17 & 68.72 \\
 & UniTok         & 7.42 & 15.50 & 10.11 & 21.04 & 12.40 & 24.87 & 14.28 & 27.45 & 16.37 & 29.48 \\
 & FSQ            & 0.51 & 0.53 & 1.00 & 1.09 & 2.16 & 2.36 & 4.77 & 4.57 & 7.23 & 7.07 \\
\midrule
\multirow{4}{*}{ISIC}
 & LlamaGen VQ-16 & 17.43 & 29.58 & 27.33 & 53.05 & 39.29 & 83.25 & 49.71 & 109.98 & 58.15 & 128.31 \\
 & MedITok        & 6.44 & 13.16 & 12.38 & 29.02 & 19.63 & 52.93 & 27.30 & 70.50 & 34.75 & 75.45 \\
 & UniTok         & 7.19 & 14.22 & 9.75 & 18.94 & 12.05 & 23.06 & 14.18 & 26.22 & 16.26 & 28.64 \\
 & FSQ            & 0.42 & 0.44 & 0.82 & 0.89 & 1.69 & 1.88 & 3.46 & 3.64 & 5.00 & 5.97 \\
\bottomrule
\end{tabular}%
}

\end{table}

\begin{table}[htbp]
\centering
\small
\setlength{\tabcolsep}{4pt}
\caption{Token-path robustness (probe AUC) under M2-unb, all datasets, full $\epsilon$ grid.
clean is the unattacked probe AUC. $\epsilon$ in the pixel
convention. MedITok collapses on every dataset. Lower is a stronger attack.
$n{=}256$ per dataset.}
\label{tab:probe_m3unb_all}
\begin{tabular}{llcccccc}
\toprule
& Tokenizer & clean & $\nicefrac{0.5}{255}$ & $\nicefrac{1}{255}$ & $\nicefrac{2}{255}$ & $\nicefrac{4}{255}$ & $\nicefrac{8}{255}$ \\
\midrule
\multirow{4}{*}{Kermany}
 & LlamaGen VQ-16 & 0.877 & 0.904 & 0.904 & 0.874 & 0.874 & 0.731 \\
 & MedITok        & 0.938 & 0.780 & 0.509 & 0.204 & 0.095 & 0.056 \\
 & UniTok         & 0.921 & 0.912 & 0.880 & 0.790 & 0.587 & 0.485 \\
 & FSQ            & 0.869 & 0.865 & 0.863 & 0.856 & 0.839 & 0.821 \\
\midrule
\multirow{4}{*}{APTOS}
 & LlamaGen VQ-16 & 0.951 & 0.946 & 0.926 & 0.879 & 0.822 & 0.734 \\
 & MedITok        & 0.987 & 0.887 & 0.570 & 0.134 & 0.054 & 0.038 \\
 & UniTok         & 0.980 & 0.964 & 0.914 & 0.791 & 0.511 & 0.398 \\
 & FSQ            & 0.954 & 0.938 & 0.891 & 0.730 & 0.547 & 0.491 \\
\midrule
\multirow{4}{*}{ISIC}
 & LlamaGen VQ-16 & 0.762 & 0.758 & 0.777 & 0.753 & 0.731 & 0.661 \\
 & MedITok        & 0.834 & 0.814 & 0.725 & 0.378 & 0.228 & 0.198 \\
 & UniTok         & 0.821 & 0.732 & 0.696 & 0.581 & 0.464 & 0.410 \\
 & FSQ            & 0.743 & 0.729 & 0.723 & 0.721 & 0.677 & 0.648 \\
\bottomrule
\end{tabular}

\end{table}

\begin{table}[htbp]
\centering
\small
\setlength{\tabcolsep}{4pt}
\caption{Reconstruction-path robustness (recon AUC) under M2-unb, all datasets, full $\epsilon$ grid.
The geometric attack barely degrades this path except at the largest
budget. $\epsilon$ in the pixel convention; lower is a stronger attack.
$n{=}256$ per dataset.}
\label{tab:recon_m3unb_all}
\begin{tabular}{llcccccc}
\toprule
& Tokenizer & clean & $\nicefrac{0.5}{255}$ & $\nicefrac{1}{255}$ & $\nicefrac{2}{255}$ & $\nicefrac{4}{255}$ & $\nicefrac{8}{255}$ \\
\midrule
\multirow{4}{*}{Kermany}
 & LlamaGen VQ-16 & 0.957 & 0.943 & 0.931 & 0.923 & 0.787 & 0.679 \\
 & MedITok        & 0.959 & 0.956 & 0.964 & 0.941 & 0.765 & 0.626 \\
 & UniTok         & 0.934 & 0.939 & 0.930 & 0.925 & 0.873 & 0.845 \\
 & FSQ            & 0.810 & 0.820 & 0.825 & 0.814 & 0.810 & 0.785 \\
\midrule
\multirow{4}{*}{APTOS}
 & LlamaGen VQ-16 & 0.997 & 0.992 & 0.989 & 0.976 & 0.943 & 0.926 \\
 & MedITok        & 0.995 & 0.993 & 0.986 & 0.978 & 0.942 & 0.880 \\
 & UniTok         & 0.993 & 0.994 & 0.991 & 0.983 & 0.967 & 0.939 \\
 & FSQ            & 0.932 & 0.919 & 0.929 & 0.915 & 0.905 & 0.883 \\
\midrule
\multirow{4}{*}{ISIC}
 & LlamaGen VQ-16 & 0.882 & 0.890 & 0.892 & 0.803 & 0.767 & 0.650 \\
 & MedITok        & 0.910 & 0.902 & 0.878 & 0.850 & 0.707 & 0.607 \\
 & UniTok         & 0.910 & 0.907 & 0.887 & 0.854 & 0.794 & 0.764 \\
 & FSQ            & 0.797 & 0.801 & 0.794 & 0.821 & 0.784 & 0.686 \\
\bottomrule
\end{tabular}

\end{table}

\FloatBarrier

\subsection{Full attack and defense sweep data}
\label{app:full-sweeps}

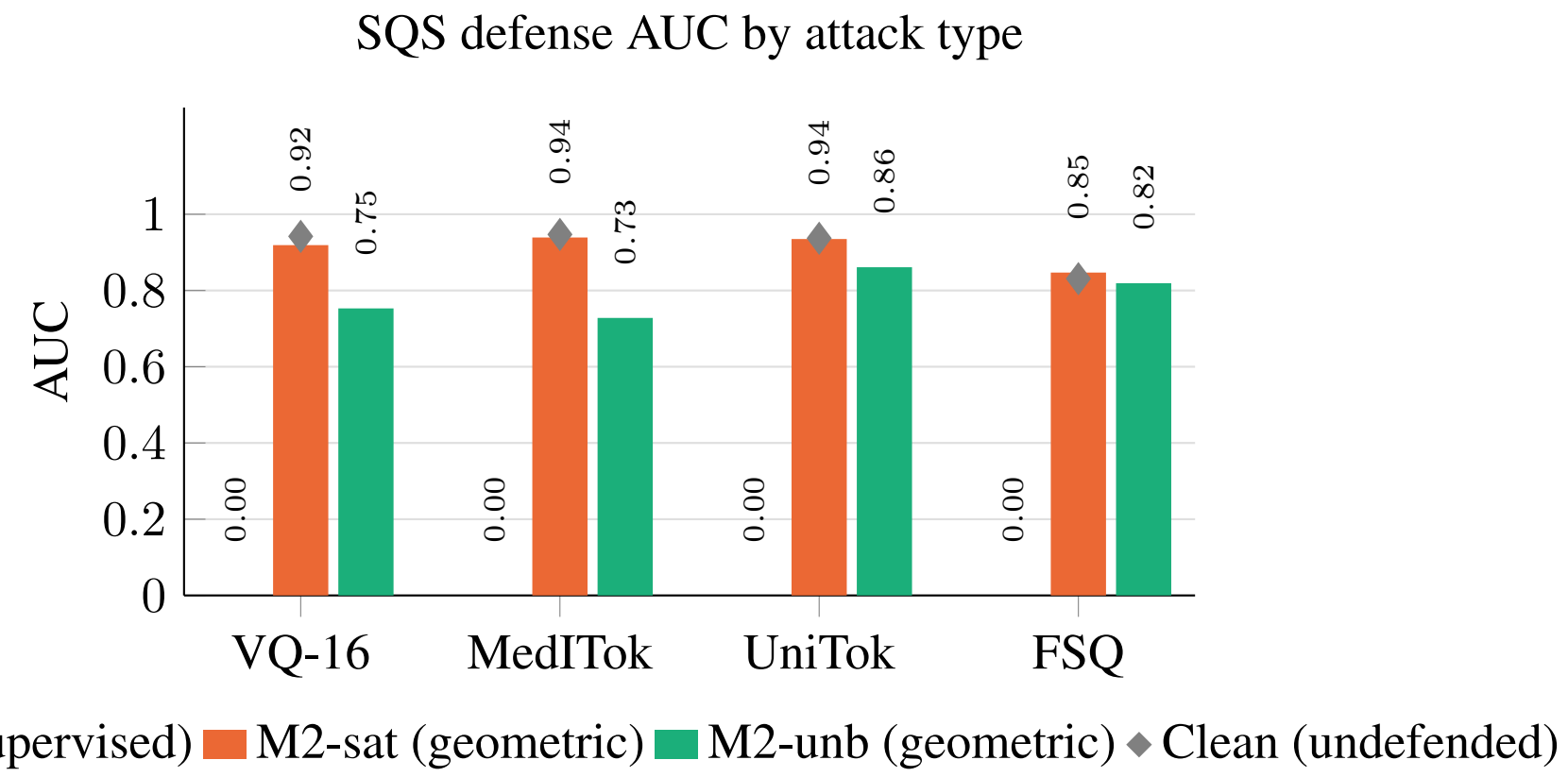
\begin{figure}[htbp]
\centering
\begin{tikzpicture}
\begin{axis}[
    ybar,
    bar width=11pt,
    width=0.62\linewidth,
    height=5.0cm,
    ymin=0, ymax=1.28, ytick={0,0.2,0.4,0.6,0.8,1},
    ylabel={AUC},
    title={SQS defense AUC by attack type},
    symbolic x coords={VQ-16,MedITok,UniTok,FSQ},
    xtick=data,
    enlarge x limits=0.15,
    axis lines*=left,
    ymajorgrids,
    grid style={gray!25},
    tick label style={font=\normalsize},
    label style={font=\normalsize},
    title style={font=\normalsize},
    legend style={at={(0.5,-0.22)},anchor=north,legend columns=-1,draw=none,font=\normalsize},
    ]
\addplot[fill=vizone, draw=none, barlegend=vizone,
    nodes near coords, nodes near coords style={font=\tiny, rotate=90, anchor=west, xshift=7pt,
        /pgf/number format/fixed, /pgf/number format/precision=2, /pgf/number format/zerofill}]
    coordinates {(VQ-16,0.000) (MedITok,0.000) (UniTok,0.000) (FSQ,0.000)};
\addplot[fill=viztwo, draw=none, barlegend=viztwo,
    nodes near coords, nodes near coords style={font=\tiny, rotate=90, anchor=west, xshift=7pt,
        /pgf/number format/fixed, /pgf/number format/precision=2, /pgf/number format/zerofill}]
    coordinates {(VQ-16,0.919) (MedITok,0.939) (UniTok,0.935) (FSQ,0.847)};
\addplot[fill=vizthree, draw=none, barlegend=vizthree,
    nodes near coords, nodes near coords style={font=\tiny, rotate=90, anchor=west, xshift=7pt,
        /pgf/number format/fixed, /pgf/number format/precision=2, /pgf/number format/zerofill}]
    coordinates {(VQ-16,0.753) (MedITok,0.728) (UniTok,0.861) (FSQ,0.819)};
\addplot[only marks, mark=diamond*, mark size=3pt, color=gray,
    legend image code/.code={\node[diamond, fill=gray, scale=0.55] at (0.2cm,0cm) {};}]
    coordinates {(VQ-16,0.942) (MedITok,0.947) (UniTok,0.937) (FSQ,0.831)};
\legend{M1 (supervised), M2-sat (geometric), M2-unb (geometric), Clean (undefended)}
\end{axis}
\end{tikzpicture}
\caption{SQS protects against boundary-directed attacks only, not supervised ones.
Adversarial AUC at $\epsilon=8/255$ under SQS's selected hyperparameters, mean over datasets, by
attack: M1 (supervised) collapses to $0.000$ on every tokenizer; M2-sat/M2-unb (boundary-directed)
retain most of their clean AUC; gray diamonds mark each tokenizer's clean (undefended, unattacked)
reconstruction-path AUC ($n{=}128$ per dataset).}
\label{fig:sqs-attack-dependence}
\end{figure}

Figure~\ref{fig:sqs-attack-dependence} breaks SQS's protection down by attack: against M1 SQS's AUC is
$0.00$ on every tokenizer, while against boundary-directed attacks (M2-sat/M2-unb) it retains
most of its clean AUC. Sweeping SQS's noise scale $\sigma$ and vote count $K$ moves mean
adversarial AUC by less than $0.01$ across the grid on every tokenizer, because that flat mean
averages across attacks SQS treats completely differently.

\begin{figure}[ht]
\centering
\begin{tikzpicture}
\begin{groupplot}[
    group style={group size=2 by 2, horizontal sep=1.3cm, vertical sep=1.5cm},
    width=4.8cm, height=4.2cm,
    xlabel={Clean recon.\ AUC}, ylabel={Adv.\ recon.\ AUC},
    label style={font=\small}, tick label style={font=\small},
    title style={font=\small},
    ymajorgrids, grid style={gray!25},
    ]
\nextgroupplot[title={VQ-16}, xmin=0.895, xmax=0.965, ymin=0.58, ymax=0.945, only marks]
\addplot[mark=*, mark size=2pt, vizone] coordinates {(0.942,0.760)(0.938,0.766)(0.914,0.660)(0.933,0.740)(0.911,0.634)(0.944,0.890)};
\addplot[mark=star, mark size=4pt, viztwo] coordinates {(0.951,0.895)};
\node[anchor=north west, font=\small] at (axis cs:0.942,0.760) {1};
\node[anchor=south west, font=\small] at (axis cs:0.938,0.766) {2};
\node[anchor=south, font=\small] at (axis cs:0.914,0.660) {3};
\node[anchor=north east, font=\small] at (axis cs:0.933,0.740) {4};
\node[anchor=north, font=\small] at (axis cs:0.911,0.634) {5};
\node[anchor=south east, font=\small] at (axis cs:0.944,0.890) {7};
\node[anchor=south west, font=\small] at (axis cs:0.951,0.895) {6};
\nextgroupplot[title={MedITok}, xmin=0.895, xmax=0.960, ymin=0.65, ymax=0.925, only marks]
\addplot[mark=*, mark size=2pt, vizone] coordinates {(0.947,0.697)(0.919,0.889)(0.942,0.719)(0.939,0.720)(0.944,0.732)(0.937,0.869)(0.936,0.870)};
\node[anchor=north west, font=\small] at (axis cs:0.947,0.697) {1};
\node[anchor=south east, font=\small] at (axis cs:0.919,0.889) {2};
\draw[gray, line width=1pt] (axis cs:0.942,0.719) -- (axis cs:0.954,0.700);
\node[anchor=west, font=\small] at (axis cs:0.954,0.700) {3};
\draw[gray, line width=1pt] (axis cs:0.939,0.720) -- (axis cs:0.925,0.703);
\node[anchor=east, font=\small] at (axis cs:0.925,0.703) {4};
\node[anchor=south west, font=\small] at (axis cs:0.944,0.732) {5};
\draw[gray, line width=1pt] (axis cs:0.937,0.869) -- (axis cs:0.920,0.847);
\node[anchor=east, font=\small] at (axis cs:0.920,0.847) {6};
\draw[gray, line width=1pt] (axis cs:0.936,0.870) -- (axis cs:0.950,0.853);
\node[anchor=west, font=\small] at (axis cs:0.950,0.853) {7};
\nextgroupplot[title={UniTok}, xmin=0.905, xmax=0.960, ymin=0.795, ymax=0.930, only marks]
\addplot[mark=*, mark size=2pt, vizone] coordinates {(0.937,0.856)(0.945,0.843)(0.933,0.847)(0.942,0.840)(0.943,0.890)};
\addplot[mark=star, mark size=4pt, viztwo] coordinates {(0.924,0.902)(0.938,0.900)};
\node[anchor=south, font=\small] at (axis cs:0.937,0.856) {1};
\node[anchor=north east, font=\small] at (axis cs:0.924,0.902) {2};
\node[anchor=north west, font=\small] at (axis cs:0.945,0.843) {3};
\node[anchor=south east, font=\small] at (axis cs:0.933,0.847) {4};
\node[anchor=south, font=\small] at (axis cs:0.942,0.840) {5};
\node[anchor=west, font=\small] at (axis cs:0.943,0.890) {6};
\node[anchor=south west, font=\small] at (axis cs:0.938,0.900) {7};
\nextgroupplot[title={FSQ}, xmin=0.70, xmax=0.97, ymin=0.70, ymax=0.925, only marks]
\addplot[mark=*, mark size=2pt, vizone] coordinates {(0.831,0.775)(0.838,0.773)(0.837,0.761)(0.778,0.756)(0.784,0.747)(0.930,0.880)(0.865,0.841)};
\node[anchor=south east, font=\small] at (axis cs:0.831,0.775) {1};
\node[anchor=north west, font=\small] at (axis cs:0.838,0.773) {2};
\node[anchor=south west, font=\small] at (axis cs:0.837,0.761) {3};
\node[anchor=south east, font=\small] at (axis cs:0.778,0.756) {4};
\node[anchor=north west, font=\small] at (axis cs:0.784,0.747) {5};
\node[anchor=south east, font=\small] at (axis cs:0.930,0.880) {6};
\node[anchor=north west, font=\small] at (axis cs:0.865,0.841) {7};
\end{groupplot}
\end{tikzpicture}
\caption{\small Clean vs.\ adversarial reconstruction-path AUC (M2-unb, $\epsilon{=}8/255$), mean over
datasets, one point per defense, faceted by tokenizer. $1{=}$None, $2{=}$Pixel-noise,
$3{=}$SQS, $4{=}$Enc-FT, $5{=}$Enc-FT+SQS, $6{=}$Dec-FT, $7{=}$Enc+Dec-FT.
Starred points (orange) are globally Pareto-optimal: UniTok/Pixel-noise, UniTok/Enc+Dec-FT,
VQ-16/Dec-FT. $n{=}128$ per dataset.}
\label{fig:defense-pareto}
\end{figure}
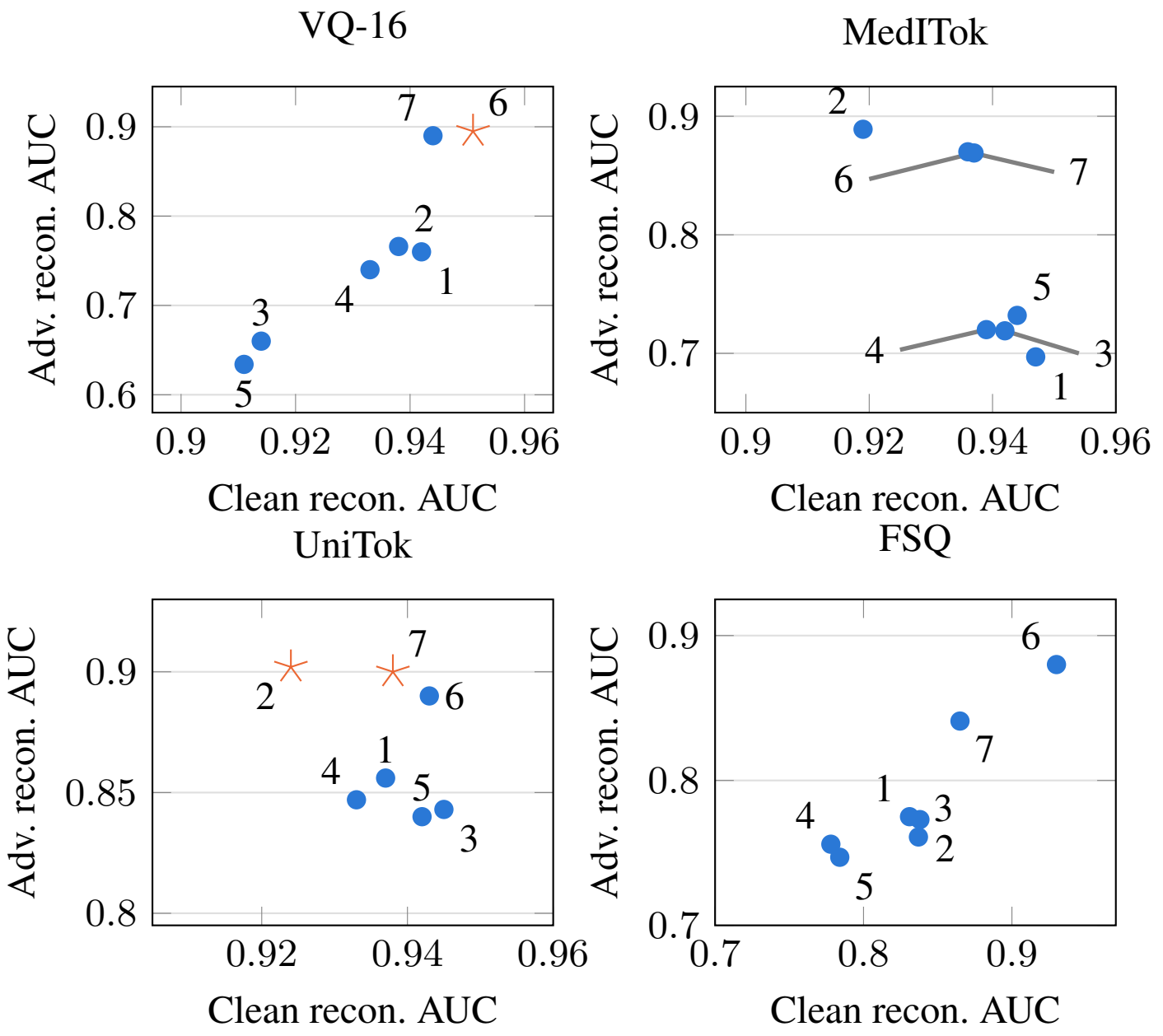

Figure~\ref{fig:defense-pareto} re-plots Table~\ref{tab:defense_recon_all}'s reconstruction-path
numbers as a clean-vs-robust scatter to expose the accuracy/robustness tradeoff directly. Three
points (UniTok with Pixel-noise, UniTok with Encoder-FT+Decoder-FT, and VQ-16 with Decoder-FT,
starred) Pareto-dominate every other tokenizer/defense combination we tested.

\FloatBarrier

\FloatBarrier

\subsubsection*{Reconstruction fidelity: full per-dataset breakdown}

Table~\ref{tab:gen_fidelity_appendix_attack} and
Table~\ref{tab:gen_fidelity_defense_appendix} are the full per-dataset,
full-$\epsilon$-grid data behind the main text's reconstruction-fidelity
paragraph (Section~\ref{sec:defenses}; attack and defense comparisons
respectively).

\begin{table}[htbp]
\centering
\scriptsize
\setlength{\tabcolsep}{3.2pt}
\renewcommand{\arraystretch}{0.85}
\caption{Per-dataset reconstruction fidelity vs.\ original, M2-sat and
M2-unb, full $\epsilon$ grid. $n{=}256$ per cell. LPIPS$\downarrow$,
PSNR$\uparrow$.}
\label{tab:gen_fidelity_appendix_attack}
\begin{tabular}{llcccccccccc}
\toprule
& & \multicolumn{2}{c}{$\epsilon{=}0.5$} & \multicolumn{2}{c}{$\epsilon{=}1$} & \multicolumn{2}{c}{$\epsilon{=}2$} & \multicolumn{2}{c}{$\epsilon{=}4$} & \multicolumn{2}{c}{$\epsilon{=}8$} \\
Tokenizer / Dataset & Attack & LPIPS & PSNR & LPIPS & PSNR & LPIPS & PSNR & LPIPS & PSNR & LPIPS & PSNR \\
\midrule
\multirow{2}{*}{VQ-16 / Kermany} & M2-sat & 0.104 & 25.47 & 0.106 & 25.40 & 0.112 & 25.30 & 0.131 & 25.13 & 0.213 & 24.72 \\
 & M2-unb & 0.103 & 25.51 & 0.108 & 25.42 & 0.138 & 25.07 & 0.277 & 24.07 & 0.485 & 22.47 \\
\multirow{2}{*}{VQ-16 / APTOS} & M2-sat & 0.072 & 30.39 & 0.077 & 30.25 & 0.091 & 29.95 & 0.119 & 29.62 & 0.200 & 29.14 \\
 & M2-unb & 0.073 & 30.37 & 0.083 & 30.09 & 0.128 & 29.62 & 0.228 & 28.57 & 0.363 & 27.32 \\
\multirow{2}{*}{VQ-16 / ISIC} & M2-sat & 0.113 & 27.96 & 0.121 & 27.81 & 0.139 & 27.62 & 0.171 & 27.36 & 0.239 & 26.92 \\
 & M2-unb & 0.116 & 27.92 & 0.133 & 27.58 & 0.192 & 27.13 & 0.318 & 26.11 & 0.473 & 24.86 \\
\midrule
\multirow{2}{*}{FSQ / Kermany} & M2-sat & 0.236 & 27.53 & 0.242 & 27.39 & 0.245 & 27.29 & 0.243 & 27.31 & 0.256 & 27.11 \\
 & M2-unb & 0.233 & 27.60 & 0.234 & 27.58 & 0.234 & 27.58 & 0.250 & 27.34 & 0.321 & 26.28 \\
\multirow{2}{*}{FSQ / APTOS} & M2-sat & 0.160 & 29.61 & 0.173 & 29.25 & 0.179 & 28.86 & 0.179 & 29.04 & 0.209 & 28.87 \\
 & M2-unb & 0.157 & 29.63 & 0.159 & 29.62 & 0.159 & 29.61 & 0.196 & 29.31 & 0.298 & 27.98 \\
\multirow{2}{*}{FSQ / ISIC} & M2-sat & 0.201 & 28.60 & 0.209 & 28.43 & 0.210 & 28.30 & 0.209 & 28.29 & 0.234 & 27.99 \\
 & M2-unb & 0.199 & 28.68 & 0.199 & 28.71 & 0.198 & 28.73 & 0.238 & 28.30 & 0.360 & 26.78 \\
\midrule
\multirow{2}{*}{MedITok / Kermany} & M2-sat & 0.037 & 31.10 & 0.039 & 31.07 & 0.043 & 31.02 & 0.066 & 30.57 & 0.158 & 29.62 \\
 & M2-unb & 0.037 & 31.25 & 0.044 & 30.71 & 0.083 & 28.67 & 0.216 & 25.30 & 0.423 & 24.08 \\
\multirow{2}{*}{MedITok / APTOS} & M2-sat & 0.028 & 35.27 & 0.030 & 35.20 & 0.037 & 34.90 & 0.062 & 34.18 & 0.148 & 32.57 \\
 & M2-unb & 0.028 & 35.42 & 0.034 & 34.93 & 0.072 & 32.00 & 0.210 & 23.28 & 0.479 & 16.26 \\
\multirow{2}{*}{MedITok / ISIC} & M2-sat & 0.053 & 31.96 & 0.056 & 31.80 & 0.065 & 31.60 & 0.094 & 31.19 & 0.184 & 30.23 \\
 & M2-unb & 0.053 & 31.96 & 0.065 & 31.49 & 0.133 & 28.42 & 0.330 & 22.40 & 0.514 & 19.62 \\
\midrule
\multirow{2}{*}{UniTok / Kermany} & M2-sat & 0.044 & 30.32 & 0.046 & 30.14 & 0.051 & 30.00 & 0.074 & 29.73 & 0.154 & 29.12 \\
 & M2-unb & 0.045 & 30.27 & 0.054 & 29.63 & 0.088 & 28.43 & 0.183 & 26.88 & 0.347 & 25.28 \\
\multirow{2}{*}{UniTok / APTOS} & M2-sat & 0.036 & 34.52 & 0.038 & 34.36 & 0.045 & 34.18 & 0.066 & 33.87 & 0.137 & 32.84 \\
 & M2-unb$^\dagger$ & 0.037 & 34.30 & 0.048 & 33.46 & 0.077 & 32.17 & 0.147 & 24.46$^\dagger$ & 0.275 & 27.95 \\
\multirow{2}{*}{UniTok / ISIC} & M2-sat & 0.056 & 31.64 & 0.060 & 31.47 & 0.067 & 31.33 & 0.090 & 30.94 & 0.165 & 30.17 \\
 & M2-unb & 0.058 & 31.52 & 0.073 & 30.86 & 0.107 & 29.86 & 0.190 & 28.52 & 0.342 & 27.11 \\
\bottomrule
\end{tabular}

\vspace{3pt}
{\scriptsize $^\dagger$UniTok/APTOS under M2-unb is non-monotonic in $\epsilon$ on
PSNR alone: it dips to $24.46$ at $\epsilon{=}4$ before partially recovering to
$27.95$ at $\epsilon{=}8$, while LPIPS and SSIM on the same row stay monotonic
($0.077\to0.147\to0.275$ and $0.853\to0.801\to0.722$). This is present in the
source data as-is, not a transcription artifact, and was independently flagged
by the source notebook's own monotonicity check; we report it rather than
smoothing it out.}
\end{table}

\begin{table}[htbp]
\centering
\scriptsize
\setlength{\tabcolsep}{3.5pt}
\renewcommand{\arraystretch}{0.85}
\caption{Per-dataset defense-fidelity comparison under M2-unb,
$\epsilon{=}8/255$. $n{=}256$ per cell. LPIPS$\downarrow$, PSNR$\uparrow$,
SSIM$\uparrow$.}
\label{tab:gen_fidelity_defense_appendix}
\begin{tabular}{llccc ccc ccc}
\toprule
& & \multicolumn{3}{c}{Kermany} & \multicolumn{3}{c}{APTOS} & \multicolumn{3}{c}{ISIC} \\
Tokenizer & Defense & LPIPS & PSNR & SSIM & LPIPS & PSNR & SSIM & LPIPS & PSNR & SSIM \\
\midrule
\multirow{5}{*}{VQ-16}
 & Clean       & 0.103 & 25.60 & 0.675 & 0.070 & 30.35 & 0.842 & 0.109 & 28.07 & 0.725 \\
 & None        & 0.485 & 22.47 & 0.417 & 0.363 & 27.32 & 0.657 & 0.473 & 24.86 & 0.503 \\
 & Encoder-FT  & 0.378 & 23.56 & 0.496 & 0.338 & 27.53 & 0.674 & 0.412 & 25.27 & 0.547 \\
 & Decoder-FT  & 0.207 & 23.47 & 0.539 & 0.171 & 28.69 & 0.794 & 0.307 & 26.08 & 0.607 \\
 & +Decoder-FT & 0.185 & 23.69 & 0.569 & 0.169 & 28.09 & 0.788 & 0.273 & 26.05 & 0.626 \\
\midrule
\multirow{5}{*}{FSQ}
 & Clean       & 0.231 & 27.65 & 0.739 & 0.153 & 29.70 & 0.821 & 0.196 & 28.72 & 0.734 \\
 & None        & 0.321 & 26.28 & 0.645 & 0.298 & 27.98 & 0.700 & 0.360 & 26.78 & 0.605 \\
 & Encoder-FT  & 0.408 & 24.67 & 0.647 & 0.278 & 23.77 & 0.741 & 0.340 & 24.93 & 0.630 \\
 & Decoder-FT  & 0.241 & 26.65 & 0.646 & 0.266 & 26.12 & 0.681 & 0.326 & 24.47 & 0.502 \\
 & +Decoder-FT & 0.302 & 25.45 & 0.646 & 0.280 & 24.16 & 0.710 & 0.332 & 23.26 & 0.526 \\
\midrule
\multirow{5}{*}{MedITok}
 & Clean       & 0.036 & 31.00 & 0.863 & 0.026 & 35.43 & 0.913 & 0.050 & 32.09 & 0.843 \\
 & None        & 0.423 & 24.08 & 0.558 & 0.479 & 16.26 & 0.478 & 0.514 & 19.62 & 0.365 \\
 & Encoder-FT  & 0.398 & 24.46 & 0.578 & 0.157$^\ddagger$ & 28.16$^\ddagger$ & 0.830$^\ddagger$ & 0.472 & 20.50 & 0.407 \\
 & Decoder-FT  & 0.169 & 26.32 & 0.710 & 0.147 & 31.13 & 0.850 & 0.272 & 26.61 & 0.685 \\
 & +Decoder-FT & 0.166 & 26.45 & 0.718 & 0.155 & 26.30 & 0.850 & 0.249 & 26.79 & 0.697 \\
\midrule
\multirow{5}{*}{UniTok}
 & Clean       & 0.042 & 30.48 & 0.832 & 0.033 & 34.63 & 0.891 & 0.053 & 31.78 & 0.817 \\
 & None        & 0.347 & 25.28 & 0.599 & 0.275 & 27.95 & 0.722 & 0.342 & 27.11 & 0.617 \\
 & Encoder-FT  & 0.233 & 27.60 & 0.700 & 0.201 & 30.51 & 0.792 & 0.263 & 27.95 & 0.657 \\
 & Decoder-FT  & 0.127 & 27.39 & 0.727 & 0.122 & 32.44 & 0.864 & 0.179 & 29.25 & 0.751 \\
 & +Decoder-FT & 0.111 & 27.72 & 0.740 & 0.109 & 32.03 & 0.871 & 0.156 & 29.23 & 0.760 \\
\bottomrule
\end{tabular}

\vspace{3pt}
{\scriptsize $^\ddagger$Two Encoder-FT checkpoints in the source data
(Kermany/UniTok's $\epsilon{=}2$ checkpoint and APTOS/MedITok's $\epsilon{=}1$
checkpoint) have fidelity metrics 10+~dB worse than their neighboring
checkpoints on every metric simultaneously and are excluded as apparently
broken checkpoints rather than genuine defense results; this table uses only
the $\epsilon{=}8$ checkpoint throughout (matching the attack's own budget),
so neither excluded checkpoint enters it. The MedITok/APTOS Encoder-FT cell
above is the $\epsilon{=}8$ checkpoint, not the excluded $\epsilon{=}1$ one.}
\end{table}

\paragraph{Scope.} This reconstruction-fidelity data covers Encoder-FT,
Decoder-FT, and Encoder-FT+Decoder-FT only; SQS and Pixel-noise were not run
through these fidelity metrics, so their effect on reconstruction
\emph{quality} (as opposed to downstream AUC, Table~\ref{tab:defense_recon_all})
is not reported. $n{=}256$ per cell throughout, with the FID reliability flag
low ($n{<}2048$) at every cell; only LPIPS/PSNR/SSIM are reported for this
reason, not FID.

%%%%%%%%
%%
% - **Format:** Main paper figures are PNG, appendix figures are JPG. Neither is vector.
%   All should be PDF. Use `matplotlib.backends.backend_pdf` or `savefig('name.pdf')`.
% - **Font size:** All axis labels and tick labels should be at least 10pt in the rendered
%   PDF. Check at 1:1 zoom in the compiled PDF, not in the raw figure file.
% - **Line thickness:** Lines should be at least 1.5pt. Thin lines disappear in print.
% - **Color palette:** Use a colorblind-safe palette (e.g., matplotlib's `tab10` or the
%   Wong palette). Verify figures are distinguishable in grayscale (dashes vs dots vs
%   solid styles in addition to color).
% - **Legend:** Legends should be inside the plot or in a consistent position. All
%   series must be labeled. Do not rely on color alone.
% - **Axes:** Every axis must have a label with units. Eps axis should state "/255" units.
%   AUC axis should be labeled "AUC" or "ROC-AUC". Flip rate should be labeled as
%   "Token flip rate (%)".
% - **Titles:** Figures should not have matplotlib-style titles (these go in the LaTeX
%   caption). The plot area should be clean.
% - **Consistent clean baseline:** Where AUC is plotted, mark the clean AUC as a
%   horizontal dashed line for reference.
% - **Figure sizing:** All main-paper figures should render at the same aspect ratio.
%   Appendix figures may be taller but should use consistent width.
  %%%%%

\newpage
%%%%%%%%%%%%%%%%%%%%%%%%%%%%%%%%%%%%%%%%%%%%%%%%%%%%%%%%%%%%
\FloatBarrier
\clearpage 

\end{document}